\documentclass[11pt,a4paper]{article}
\usepackage{adjustbox}
\usepackage{amsmath}
\usepackage{algorithm}
\usepackage{algorithmicx}
\usepackage{bm}
\usepackage{bbm}
\usepackage{caption}
\usepackage[T1]{fontenc}
\usepackage{graphicx}
\usepackage{subfig}
\usepackage{optidef}

\usepackage{times,latexsym}
\usepackage{url}

\usepackage[acceptedWithA]{tacl2021v1}

\usepackage{xspace,mfirstuc,tabulary}

\newif\iftaclinstructions
\taclinstructionsfalse 
\iftaclinstructions
\renewcommand{\confidential}{}
\renewcommand{\anonsubtext}{(No author info supplied here, for consistency with
TACL-submission anonymization requirements)}
\newcommand{\instr}
\fi

\iftaclpubformat 

\else

\fi

\newcommand{\com}[1]{}

\newcolumntype{\vthick}{!{\vrule width 1.5pt}}

\newcommand{\guyfix}[1]{\textcolor{black}{#1}}

\title{Multi-task Active Learning for Pre-trained Transformer-based Models}

\author{
Guy Rotman \and Roi Reichart\\
Faculty of Industrial Engineering and Management, Technion, IIT\\
\texttt{grotman@campus.technion.ac.il} \\
\hspace{1.2em}\texttt{roiri@technion.ac.il} \\
}

\date{}

\begin{document}
\maketitle
\begin{abstract}

Multi-task learning, in which several tasks are jointly learned by a single model, allows NLP models to share information from multiple annotations and may facilitate better predictions when the tasks are inter-related. This technique, however, requires annotating the same text with multiple annotation schemes which may be costly and laborious.
Active learning (AL) has been demonstrated to optimize annotation processes by iteratively selecting unlabeled examples whose annotation is most valuable for the NLP model. Yet, multi-task active learning (MT-AL) has not been applied to state-of-the-art pre-trained Transformer-based NLP models. This paper aims to close this gap. We explore various multi-task selection criteria in three realistic multi-task scenarios, reflecting different relations between the participating tasks, and demonstrate the effectiveness of multi-task compared to single-task selection. Our results suggest that MT-AL can be effectively used in order to minimize annotation efforts for multi-task NLP models.\footnote{Our code base is available at: \url{https://github.com/rotmanguy/MTAL}.} \footnote{Accepted for publication in Transactions of the Association for Computational Linguistics (TACL), 2022. Pre-MIT Press publication version
}
\end{abstract}

\section{Introduction}
\label{sec:intro}
Deep neural networks (DNNs) have recently achieved state-of-the-art results for many natural language processing (NLP) tasks and applications. 
Of particular importance are contextualized embedding models \cite{mccann2017learned,peters2018deep}, most of which implement transformer-based architectures with the self-attention mechanism \cite{vaswani2017attention, devlin2019bert, raffel2020exploring}.

\com{
With the rise of DNNs, classical statistical methods such as PPMI and TF-IDF, have been pushed aside by their neural variants such as glove and W2vec, skip-vectors. These neural models have been proven more effective for representing textual units, especially words. However, such models are constrained to constitute only a single vector for every word in the vocabulary, while many words often symbolize multiple meanings. Lately, neural contextualized embedding models have been proposed, allowing to represent each word given its direct contextual surroundings \cite{mccann2017learned,peters2018deep}. A prominent example of such models are Transformer-based models, integrated with the self-attention mechanism, which currently hold SOTA models for many NLP applications \cite{vaswani2017attention, devlin2019bert, raffel2020exploring}.
}

Nevertheless, DNNs often require large labeled training sets in order to achieve good performance. While annotating such training sets is costly and laborious, the active learning (AL) paradigm aims to minimize these costs by iteratively selecting valuable training examples for annotation. Recently, AL has been shown effective for DNNs across various NLP tasks \cite{duong-etal-2018-active, peris-casacuberta-2018-active, ein-dor-etal-2020-active}.


An appealing capability of DNNs is performing multi-task learning (MTL): Learning multiple tasks by a single model \cite{ruder2017overview}. This stems from their architectural flexibility -- constructing increasingly deeper and wider architectures from basic building blocks -- and in their gradient-based optimization which allows them to jointly update parameters from multiple task-based objectives. Indeed, MTL has become ubiquitous in NLP \cite{luan-etal-2018-multi,liu-etal-2019-multi-task}.

MTL models for NLP can often benefit from using corpora annotated for multiple tasks, particularly when these tasks are closely-related and can inform each other. Prominent examples of multi-task corpora include OntoNotes \cite{hovy2006ontonotes}, the Universal Dependencies Bank \cite{nivre2020universal} and STREUSLE \cite{schneider2018comprehensive}. Given the importance of multi-task corpora for many MTL setups, effective AL frameworks that support MTL are becoming crucial.


Unfortunately, most AL methods do not support annotations for more than one task. Multi-task AL (MT-AL) was proposed by \citet{reichart2008multi} before the neural era, and adapted by \citet{ikhwantri2018multi} to a neural architecture. Recently, \citet{zhu-etal-2020-multitask} proposed an MT-AL model for slot filling and intent detection, focusing mostly on LSTMs \cite{hochreiter1997long}. 


In this paper, we are the first to systematically explore MT-AL for large pre-trained Transformer models. Naturally, our focus is on closely-related NLP tasks, for which multi-task annotation of the same corpus is likely to be of benefit. Particularly, we consider three challenging real-life multi-task scenarios, reflecting different relations between the participating NLP tasks: 1. Complementing tasks, where each task may provide valuable information to the other task: Dependency parsing (DP) and named entity recognition (NER); 2. Hierarchically-related tasks, where one of the tasks depends on the output of the other: Relation extraction (RE) and NER; and 3. Tasks with different annotation granularity: Slot filling (SF, token level) and intent detection (ID, sentence level). We propose various novel MT-AL methods and tailor them to the specific properties of the scenarios, in order to properly address the underlying relations between the participating tasks. Our experimental results highlight a large number of patterns that can guide NLP researchers when annotating corpora with multiple annotation schemes using the AL paradigm.



\com{
Finally. our experiments reveal that similarly to single-task learning (STL) \cite{desai2020calibration,kong-etal-2020-calibrated}, multi-task Transformer models also tend to be overconfident in their predictions. This is especially harmful to AL methods that typically base their example selection on uncertainty measures \cite{li2006confidence}: When the base model is mostly overconfident such AL methods lack information regarding which examples are likely to be most useful for its subsequent training. We hence explore, to the best of our knowledge for the first time, \textit{overconfidence reduction (a.k.a. model calibration)} in the context of MT-AL, and show that it can be achieved using the label smoothing (LS) objective \cite{szegedy2016rethinking}. Our results also indicate that by mitigating this problem, LS yields better AL performance.}

\com{
Finally, previous MT-AL works did not address out-of-distribution generalization. Ideally, AL, which aims to minimize annotation costs, should aim to yield models that can be applied to new domains without additional data annotation costs.
%
We hence analyze the effectiveness of our AL methods on cross-domain performance. One interesting result reveals that the LS objective, which reduces the overconfidence of the base model, also improves cross-domain generalization.}



\com{
Finally, MT-AL carries the opportunity to test cross-task performance: basing the selection process on the prediction of a subset of the tasks and testing performance on tasks that have not participated in the selection process. Such an opportunity can shed light on the impact of each of the tasks on the others and potentially save annotation costs and reduce inference time. Despite those benefits, we are not aware of such discussion in previous works.
}

\com{
In this paper, we focus on performing extensive experiments on BERT and exploring its ability to perform MTL for AL. We introduce a unified model for learning multiple tasks with BERT and explore multiple AL selection methods for MTL. We further analyze cross-task selection impact when predictions of the evaluated task do not participate during the AL selection process.

Despite BERT's general success, our initial experiments suggest that BERT trained with the standard cross-entropy objective tends to output overconfident predictions. Such behavior is mostly unwelcome as standard AL methods are based on uncertainty measures.
Recently, \citet{szegedy2016rethinking} introduced the label smoothing objective (LS, \cite{szegedy2016rethinking}), a technique that lowers the probability of the correct label by some factor, and allows the model to randomly attend to all other labels. In this paper, we evaluate for the first time the impact of LS on multi-task AL and show that it is able to reduce model's overconfidence, as well as to better generalize on in-domain and out-of-domain distribution settings.}

\com{
To summarize our contributions we devise the following research questions:
\vspace{-5pt}
\begin{enumerate}
\setlength{\itemsep}{-4pt}
    \item Which AL selection methods are useful when performing MTL?
    \item Can MTL improve AL performance upon single-task learning (STL)?
    \item How does cross-task selection affect MT-AL performance?
    \item Can LS mitigate the overconfidence of Transformer-based models for AL?
    \item Can LS improve the performance of MT-AL models under in-domain and out-of-domain distributions?
\end{enumerate}
}

\section{Previous Work}
\label{sec:prev}
This paper addresses a previously unexplored problem: multi-task AL (MT-AL) for NLP with pre-trained Transformer-based models. We hence start by covering AL in NLP and then proceed with multi-task learning (MTL) in NLP.

\subsection{Active Learning in NLP}

AL has been successfully applied to a variety of NLP tasks, including semantic parsing \cite{duong-etal-2018-active}, syntactic parsing \cite{reichart2009sample,li2016active}, co-reference resolution \cite{li-etal-2020-active-learning}, named entity recognition \cite{shen-etal-2017-deep} and machine translation \cite{haffari2009active}, to name a few. \com{Some works studied the connection between AL and domain adaptation \cite{rai2010domain, peshterliev2019active}.} Recent works demonstrated that models like BERT can benefit from AL in low-resource settings \cite{ein-dor-etal-2020-active, griesshaber-etal-2020-fine}, and \citet{bai2020pre} suggested basing the AL selection criterion on linguistic knowledge captured by BERT. Other works performed cost-sensitive AL, where instances may have different costs \cite{tomanek2010comparison, xie-etal-2018-cost}. However, most previous works did not apply AL for MTL, which is our main focus.

\com{
\subsection{Model Confidence in Active Learning}

Model confidence is a prominent example selection criterion in AL \cite{li2006confidence, dredze2008active}. According to the confidence criterion, unlabeled examples with the lowest model confidence should be selected for annotation. Ideally, for uncertainty-based AL to work well, we would like our model to be \textit{calibrated}, that is, the correlation between its confidence and the actual quality of its output should be as high as possible \cite{beven1992future}.

Some works studied the connection between model confidence and AL. \citet{zhu2010confidence} introduced four confidence-based stopping criteria for overconfidence reduction and improved AL. \citet{mund2015active} reduced overconfidence for object classification by increasing the training importance of likely misclassified examples. Recently, \citet{chang2019bias} proposed policy reinforcement learning to alternate between confidence and other AL selection strategies.}

\com{Lastly, \citet{desai2020calibration} and \citet{kong-etal-2020-calibrated} demonstrated that label smoothing (LS) \cite{szegedy2016rethinking} can decrease calibration errors of Transformer models. In this paper, we are the first to address the impact of model calibration on Transformer-based AL systems. We show that LS achieves better calibration errors than strong baselines and leads to better performance.}

\subsection{Multi-task Learning (MTL) in NLP}

MTL has become increasingly popular in NLP, particularly when the solved tasks are closely-related \cite{chen-etal-2018-joint, safi-samghabadi-etal-2020-aggression, zhao-etal-2020-spanmlt}. In some cases, the MTL model is trained in a hierarchical fashion, where information is propagated from lower-level (sometimes auxiliary) tasks to higher-level tasks \cite{sogaard2016deep, rotman2019deep, sanh2019hierarchical, wiatrak-iso-sipila-2020-simple}. In other cases, different labeled corpora can be merged to serve as multi-task benchmarks \cite{mccann2017learned, wang-etal-2018-glue}. This way, a single MTL model can be trained on multiple tasks, which are typically only distantly-related. This research considers the setup of closely-related tasks where annotating a single corpus w.r.t. multiple tasks is a useful strategy. 

\com{
In recent years, with the rise of DNNs, much focus has been given to jointly solving multiple NLP tasks by a single model. Various model architectures have been proposed, including hard parameter sharing \cite{caruana1993multi-task,lin2018multi}, where a subset of the model parameters are shared across tasks, and soft parameter sharing \cite{duong2015low}, where the values of task-specific parameters are jointly constrained. Additional methods include adaptive dynamic sharing \cite{zaremoodi-etal-2018-adaptive} and sluice networks \cite{ruder2019latent}. The MTL architecture we consider in our experiments consists of cross-task shared modules, alongside task-specific ones. We treat all tasks as equally important during training, giving their objectives equal weight.}

\com{
When performing MTL, each of the learned tasks can have different importance. At times, we are interested in solving a particular task, referred to as the main task, while other related tasks, referred to as the auxiliary tasks, are added to allow the model to better learn the main task of interest. \cite{louvan-magnini-2018-exploring,rotman2019deep,schroder-biemann-2020-estimating}. Often, the auxiliary task is placed in lower layers, such that the learning is performed hierarchically \cite{sanh2019hierarchical, nguyen-nguyen-2021-phonlp}. Other works assume that all tasks have the same importance, for example when translating from a single source language to multiple target languages \cite{dong-etal-2015-multi} or when tasks are similar in nature \cite{safi-samghabadi-etal-2020-aggression}. In this work, we explore the MTL paradigm when performing AL given that all tasks are equally important.}

\com{
\subsection{Multi-task Learning for Active Learning}
MTL for AL has been firstly introduced by \citet{reichart2008multi} for non-neural models. The authors trained separately a named entity classifier and a dependency parser and alternated the AL selection between the two models. The alternated selection technique was then adapted by \citet{ikhwantri2018multi} to neural model based on the LSTM-CRF architecture \cite{huang2015bidirectional}. Results were reported on the semantic role labeling task, where prediction of named entity roles served as the auxiliary task. \citet{zhu-etal-2020-multi-task} also used the CRF decoder to model the shared probability distribution of the two trained tasks. Most results were conducted using the LSTM-CRF architecture, while additional results using the BERT encoder were limited. Recently, \citet{zhou2021mtaal} built an MT-AL system for named entity recognition and normalization, basing their architecture on hard-parameter sharing using an LSTM. AL selections were based on confidence summation over the two tasks. 

In this paper, we are the first to conduct extensive experiments of MT-AL over 6 different English domains with BERT, a Transformer-based language encoder. In addition to proposing new AL selections for MTL, we also examine the effects of cross-task selection, i.e., when selecting samples according to one of the tasks and evaluating them on the other. Moreover, we firstly draw the connection between model confidence to MT-AL performance, both on in-domain and on cross-domain settings.
}
\section{Task Definition - Multi-task Active Learning (MT-AL)}
\label{sec:method}

\begin{algorithm}[b!]
\caption{\footnotesize Multi-task Confidence-based Active Learning (Confidence-based MT-AL) }\label{alg:confidence_base_active_learning}
{\footnotesize
\textbf{Input:} Labeled data $\mathbf{L}$ (annotated on $t$ tasks), Unlabeled data $\mathbf{U}$\\
$\textbf{Algorithm:}$ \\
For $i=1,\ldots, \mathcal{T}$:
\begin{enumerate}
    \item Train a multi-task learning (MTL) model $h$ on $\mathbf{L}$.
    \item For each $u \in \mathbf{U}$ calculate its aggregated confidence score $C_{h}(u)$ on all $t$ tasks according to $h$.
    \item Choose the $n_i$ unlabeled examples from $\mathbf{U}$ with the lowest confidence score $C_{h}(u)$ and send them for annotation according to all $t$ tasks.
    \item Add the newly labeled examples to $\mathbf{L}$ and remove them from $\mathbf{U}$.
\end{enumerate}
}
\end{algorithm}

In the MT-AL setup, the AL algorithm is provided with a textual corpus, where an initial (typically small) set of $n_0$ examples is labeled for $t$ tasks. The AL algorithm implements an iterative process, where at the $i$-th iteration the goal of the AL algorithm is to select $n_i$ additional unlabeled examples that will be annotated on all $t$ tasks, such that the performance of the base NLP model will be improved as much as possible with respect to all of them. While such a greedy strategy of gaining the most in the $i$-th iteration may not yield the best performance in subsequent iterations, most AL algorithms are greedy, and we hence follow this strategy here as well. 

We focus on the standard setup of confidence-based AL, where unlabeled examples with the lowest model confidence are selected for annotation. 
Algorithm \ref{alg:confidence_base_active_learning} presents a general sketch of such AL algorithms, in the context of MTL. This framework, first introduced by \citet{reichart2008multi}, is a simple generalization of the single-task AL (ST-AL) framework, which supports the annotation of data with respect to multiple tasks. 

As discussed in \S \ref{sec:intro}, we explore several variations of the MT-AL setup: Independent tasks that inform each other (\S \ref{sec:closely_related_tasks}), hierarchically-related tasks, where one task depends on the output of the other (\S \ref{sec:hierarchical_tasks}), and tasks with different annotation granularity, word and sentence level (\S \ref{sec:cost_sensitive_results}). Before we can introduce the MT-AL algorithms for each of these setups, we first need to lay their shared foundations: The single-task and multi-task model confidence scores.


\section{Confidence Estimation in Single-task and Multi-task Active Learning}
\label{sec:al_methods}

We now introduce the confidence scores that we consider for single-task (ST-AL) and multi-task (MT-AL) active learning. These confidence scores are essentially the core of confidence-based AL algorithms (see Steps 2-3 of Algorithm \ref{alg:confidence_base_active_learning}). \guyfix{In Table \ref{table:selection_methods_summary} we provide a summary of the various ST-AL and MT-AL selection methods we explore.}


\subsection{Single-task Confidence Scores}

We consider three confidence scores that have been widely used in ST-AL:\\
\textbf{Random (\textit{ST-R})} This baseline method simply assigns random scores to the unlabeled examples.
\textbf{Entropy-based Confidence (\textit{ST-EC})} The single-task entropy-based confidence score is defined as:
\begin{equation}
    \textit{ST-EC}(x) = 1 - \textit{E}(x).
\end{equation}

For sentence classification tasks such as ID, \textit{E}(x) is simply the entropy over the class predictions of a sample $x$ divided by the log number of labels. In our token classification tasks (DP, NER, RE and SF), \textit{E}(x) is the normalized sentence-level entropy \cite{kim2006mmr}, which allows us to estimate the uncertainty of the model for a given sequence of tokens $x = (x_1\ldots x_m)$:
\begin{equation}
    \textit{E}(x) = - \frac{1}{m \cdot \log{s}} \sum_{i=1}^{m} \sum_{j=1}^{s} p(y_j|x_i) \log{p(y_j|x_i}),
\end{equation}
where $m$ is the number of tokens, $y_j$ is the $j$'th possible label, and $s$ is the number of labels. We perform entropy normalization by averaging the token-level entropies, in order to mitigate the effect of the sentence length, and by dividing the score by the log number of labels. The resulting confidence score ranges from 0 to 1, where lower values indicate lower certainty.
%

\textbf{Dropout Agreement (\textit{ST-DA})} Ensemble methods have proven effective for AL (see, e.g., \cite{seung1992query,settles2008analysis}) \com{\cite{seung1992query,settles2008analysis,majidi2013active}}. In this paper, we derive a confidence score inspired by \citet{reichart2007ensemble}. We start by creating $k=10$ different models by performing \textit{dropout inference} for $k$ times \cite{gal2016dropout}. We then compute the single-task dropout agreement score for a sentence $x$ by calculating the average token-level agreement across model pairs:
\begin{equation}
    \textit{ST-DA}(x) = \frac{1}{m \cdot k\cdot (k-1)} \sum_{j\neq j'} \sum_{i=1}^{m} \mathbbm{1}_{\{\hat{y}_i^{j} = \hat{y}_i^{j'}\}},
\end{equation}
where $\hat{y}_i^{j}$ is the predicted label of model $j$ for the $i$'th token.
The resulting scores range from 0 to 1, where lower values indicate lower certainty.\footnote{For sentence classification tasks, \textit{ST-DA} is computed similarly, without averaging over the tokens.}

\com{
\subsection{Multi-task Confidence Scores}
The multi-task AL (MT-AL) methods operate over multi-task models. Here, we consider seven such methods: 

\textbf{Random} Simply samples unlabeled samples at random given the predictions of the multi-task model and is denoted as multi-task random (\textit{MT-R}).

\textbf{Confidence Entropy}
Similarly to \textit{ST-EC}, we define the multi-task confidence entropy (\textit{MT-EC}) by calculating the confidence entropy (EC) score over the multi-task model.
Since the multi-task model is trained over multiple tasks, we can choose to compute the confidence-entropy scores according to samples of either of the tasks. This selection method enables us to estimate the impact of the multi-task model for each task individually when the EC criterion is calculated on the same task as that trained by the single-task model. Furthermore, we can measure the impact of the cross-task confidence selection when the evaluated task is different from the task used for the computation of the EC criterion.

\textbf{Dropout Agreement}
Similarly to \textit{ST-DA}, we define the multi-task dropout agreement (\textit{MT-DA}) by calculating the dropout agreement score over the multi-task model. Again, the \textit{MT-DA} criterion can be calculated for either of the tasks.

\textbf{Joint Confidence Entropy}
The multi-task joint confidence entropy (\textit{MT-AVG}) selection method sums for a given sample $x$ the confidence scores of all tasks:
\begin{equation}
    \textit{MT-AVG}(x) = \frac{1}{t} \sum_{i=1}^t \textit{MT-EC-i}(x),
\end{equation}
where $\textit{MT-EC-i}(x)$ represents the EC score of the \textit{i}'th task for $x$ calculated by the multi-task model.

\textbf{Joint Dropout Agreement}
Similarly to \textit{MT-AVG}, we define the multi-task joint dropout agreement (\textit{MT-AVGDA}) by replacing $\textit{MT-EC-i}(x)$ with $\textit{MT-DA-i}(x)$.

\textbf{Joint Max Confidence Entropy}
Similarly to \textit{MT-AVG}, we define the multi-task joint maximum confidence entropy (\textit{MT-MAX}) by performing the max operation, instead of summation, over the entropies of the different tasks:
\begin{equation}
    \textit{MT-MAX}(x) = \max \{\textit{MT-EC-i}(x)\}_{i=1}^t.
\end{equation}

\textbf{Joint Min Confidence Entropy}
Similarly to \textit{MT-MAX}, we define the multi-task joint minimum confidence entropy (\textit{MT-MIN}) by replacing the $\max$ operator with the $\min$ operator.
}

\subsection{Multi-task Confidence Scores}
\label{sec:mt-al-confidence}

When deriving confidence scores for MT-AL, multiple design choices should be made. First, the confidence score of a multi-task model can be based on both tasks or only on one of them. We denote with \textbf{\textit{MT-EC}} and \textbf{\textit{MT-DA}} the confidence scores that are equivalent to \textit{ST-EC} and \textit{ST-DA}: The only (important) difference is that they are calculated for a multi-task model. For clarity, we will augment this notation with the name of the task according to which the confidence is calculated. For example, \textit{ST-EC-NER} and \textit{MT-EC-NER} are the EC scores calculated using the named entity recognition (NER) classifier of a single-task and a multi-task model, respectively. 

We can hence evaluate MT-AL algorithms on cross-task selection, i.e., when the evaluated task is different from the task used for computing the confidence scores (and hence for sample selection). For example, evaluating the performance of a multi-task model, trained jointly on NER and DP, on the DP task when the confidence scores used by the MT-AL algorithm are only based on the NER classifier (\textit{MT-EC-NER}).

We also consider a family of confidence scores for MT-AL that are computed with respect to all participating tasks (joint-selection scores). For this aim, we consider three simple aggregation schemes using the average, maximum, or minimum operators over the single-task confidence scores. For example, the multi-task average confidence (\textbf{\textit{MT-AVG}}) averages for a sample $x$ the entropy-based confidence scores over all $t$ tasks:
\begin{equation}
    \textit{MT-AVG}(x) = \frac{1}{t} \sum_{i=1}^t \textit{MT-EC-i}(x),
\end{equation}

The multi-task average dropout agreement score (\textbf{\textit{MT-AVGDA}}) is similarly defined, but the averaging is over the \textit{MT-DA} scores. Finally, the multi-task maximum (minimum) \textbf{\textit{MT-MAX}} (\textbf{\textit{MT-MIN}}) is computed in a similar manner to \textit{MT-AVG} but with the $\max$ ($\min$) operator taken over the task-specific confidence entropies. 

\begin{table*}[!ht]
\centering
\begin{adjustbox}{width=\textwidth}
\begin{tabular}{|l|l|l|l|}
\hline
Selection Method & Description                                                & \begin{tabular}[c]{@{}c@{}} Participating Tasks\\ for Training\end{tabular} & \begin{tabular}[c]{@{}c@{}} Participating Tasks\\ for Selection\end{tabular} \\ \hline
\textit{ST-R}             & Single-task random selection                               & One                                                                         & None                                                                          \\ \hline
\textit{ST-EC}            & Single-task entropy-based confidence                       & One                                                                         & One                                                                          \\ \hline
\textit{ST-DA}           & Single-task dropout agreement                              & One                                                                         & One                                                                          \\ \hline
\textit{MT-R}             & Multi-task random selection                                & All                                                                         & None                                                                          \\ \hline
\textit{MT-EC}            & Multi-task entropy-based confidence                        & All                                                                         & One                                                                          \\ \hline
\textit{MT-DA}            & Multi-task dropout agreement                               & All                                                                         & One                                                                          \\ \hline
\textit{MT-AVG}           & Multi-task average entropy-based confidence                & All                                                                         & All                                                                          \\ \hline
\textit{MT-AVGDA}         & Multi-task average dropout agreement                       & All                                                                         & All                                                                          \\ \hline
\textit{MT-MAX}           & Multi-task maximum entropy-based confidence                & All                                                                         & All                                                                          \\ \hline
\textit{MT-MIN}           & Multi-task minimum entropy-based confidence                & All                                                                         & All                                                                          \\ \hline
\textit{MT-PAR}           & Multi-task Pareto entropy-based confidence                 & All                                                                         & All                                                                          \\ \hline
\textit{MT-RRF}           & Multi-task Reciprocal Rank Fusion entropy-based confidence & All                                                                         & All                                                                          \\ \hline
\textit{MT-IND}           & Multi-task independent selection entropy-based confidence  & All                                                                         & All                                                                          \\ \hline
\end{tabular}
\end{adjustbox}
\caption{Summary of the ST-AL and MT-AL selection methods explored in this paper.}
\label{table:selection_methods_summary}
\end{table*}

\paragraph{Beyond Direct Manipulations of Confidence Scores}

Since our focus in this paper is on multi-task selection, we would like to consider additional selection methods which go beyond the simple methods in previous work. The common principle of these methods is that they are less sensitive to the actual values of the confidence scores and instead consider the relative importance of the example to the participating tasks.

First, we consider \textbf{\textit{MT-PAR}}, which is based on the Pareto-efficient frontier \cite{lotov2008visualizing}. We start by representing each unlabeled sample as a $t$-dimensional space vector $c$, where $c_i=\textit{ST-EC-i}$ is the ST confidence score for task \textit{i}. Next, we select all samples for which the corresponding vector belongs to the Pareto-efficient frontier. A point belongs to the frontier if for every other vector $c'$ the following holds: 1. $\forall i\in [t], c_i \leq c'_i$ and 2. $\exists i\in [t], c_i < c'_i$. If the number of samples in the frontier is smaller than the total number of samples to select ($n)$, we re-iterate the procedure by removing the vectors of the selected samples and calculating the next Pareto points. If there are still $p$ points to be selected but the number of the final Pareto points ($f$) exceeds $p$, we select every $\lfloor{\frac{f}{p}}\rfloor$ point, ordered by the first axis.


Next, inspired by the field of \textit{information retrieval} we propose \textbf{\textit{MT-RRF}}. 
\guyfix{This method allows us to consider the rank of each example with respect to the participating tasks, rather than the actual confidence values. We first calculate $r_i$, the ranked list of the $i$-th task, by ranking the examples according to their \textit{ST-EC-i} scores, from lowest to highest. We next fuse the resulting $t$ ranked lists into a single ranked list $R$, using the reciprocal rank fusion (RRF) technique \cite{cormack2009reciprocal}. The RRF score of an example $x$ is computed as:
\begin{equation}
    RRF\text{-}Score(x) = \sum_{i=1}^{t}\frac{1}{k+r_i(x)},
\end{equation}
where $k$ is a constant, set to $60$, as in the original paper. The final ranking is computed over the RRF scores of the examples -- from highest to lowest. Higher-ranked examples are chosen first for annotation as they have lower confidence scores.} Finally, \textbf{\textit{MT-IND}} independently selects the $\lfloor{\frac{n}{t}}\rfloor$ most uncertain samples according to each task by ranking the \textit{ST-EC} scores and re-iterating if overlaps occur.

\begin{figure}[!t]
    \centering
\includegraphics[width=\linewidth]{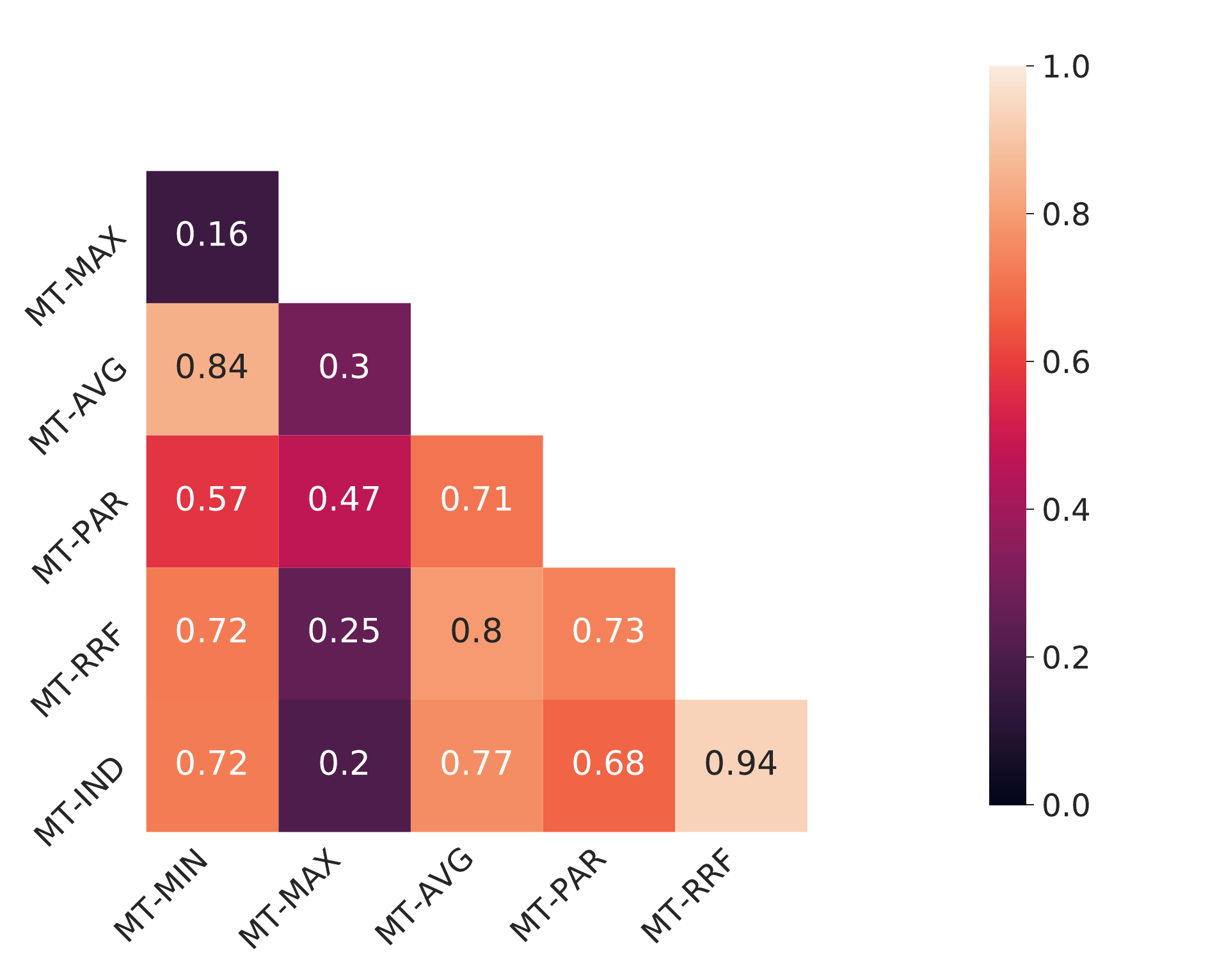}
    \caption{The percentage of shared selected samples between pairs of MT-AL selection methods (see experimental details in the text).} 
    \label{fig:mt_al_selection_methods}
\end{figure}

\com{
\begin{figure*}[!t]
    \centering
\includegraphics[width=0.99\linewidth,height=6.5cm]{figures/selection_methods_grid.png}
    \caption{An example of the chosen samples according to different MT-AL selection methods.} 
    \label{fig:mt_al_selection_methods}
\end{figure*}
}

We finally compare the selected samples of six of the MT-AL methods, after training a multi-task model for a single AL iteration on the DP and NER tasks (Figure \ref{fig:mt_al_selection_methods}). It turns out that while some of the methods tend to choose very similar example subsets for annotation (e.g., \textit{MT-IND} and \textit{MT-RRF} share 94\% of the selected samples, and \textit{MT-AVG} and \textit{MT-MIN} share 84\% of them), others substantially differ in their selection (e.g., \textit{MT-MAX} shares only 16\% of its selected samples with \textit{MT-MIN} and 20\% with \textit{MT-IND}). This observation encourages us to continue investigating the impact of the various selection methods on MT-AL.

\com{
\subsection{Limitations of the Cross-entropy Objective in Active Learning}
\label{sec:loss_funcs}

Originally, we trained our models with the standard cross-entropy (CE) loss. However, our early experiments suggested that such CE-based training yields overconfident models, which, as noted above, is likely to severely harm confidence-based AL methods. 

This phenomenon is demonstrated in Figure \ref{fig:acc_vs_conf_ce}, which presents sentence-level confidence scores as a function of sentence-level accuracy when separately training a single-task BERT-base model on DP (left figure) and on NER (right figure) with the CE objective. The confidence scores in the figure were computed according to the single-task entropy-based confidence metric ($\textit{ST-EC}$, $\S$ \ref{sec:al_methods}). The figure confirms that the model tends to be overconfident in its predictions. Furthermore, the low $R^2$ values (0.1 and 0.13 for DP and NER, respectively) indicate poor model calibration, since confidence scores are not correlated with actual accuracy. Similar patterns were observed when training our MTL model with the CE objective.

Recently, label smoothing (LS) was proposed to mitigate model overconfidence \cite{szegedy2016rethinking, muller2019when}. LS proposes to deduct a non-negative probability mass constant, $\alpha < 1$, from the probability of the correct label and to share the reserved probability mass between all labels. Formally, instead of the original CE objective $H(y,p)$, the model is trained to output a label probability distribution $p$ which minimizes $H(y^{LS},p)$, where $y^{LS}_{i} = y_i \cdot (1-\alpha) + \frac{\alpha}{|\mathcal{Y}|}$.
As noted above, this paper investigates the potential of LS to reduce the overconfidence of STL and MTL models and consequently improve AL performance. In $\S$ \ref{sec:results} (Q1.4) we compare LS with the standard CE objective, as well as with other standard calibration techniques, and show that LS is the most effective method to mitigate overconfident predictions.}

\com{
\begin{table}[!ht]
\scalebox{0.63}{
\begin{tabular}{|c|cccc|cccc|}
\hline
            & \multicolumn{4}{c|}{Default Training}                                                                                           & \multicolumn{4}{c|}{Active Learning}                                                                                         \\ \hline
            & \multicolumn{2}{c|}{DP}                                                 & \multicolumn{2}{c|}{NER}                           & \multicolumn{2}{c|}{DP}                                                 & \multicolumn{2}{c|}{NER}                           \\ \hline
            & \multicolumn{1}{c|}{Avg}            & \multicolumn{1}{c|}{Best}         & \multicolumn{1}{c|}{Avg}            & Best         & \multicolumn{1}{c|}{Avg}            & \multicolumn{1}{c|}{Best}         & \multicolumn{1}{c|}{Avg}            & Best         \\ \hline
ST          & \multicolumn{1}{c|}{87.64}          & \multicolumn{1}{c|}{0/6}          & \multicolumn{1}{c|}{71.31}          & 1/6          & \multicolumn{1}{c|}{88.96}          & \multicolumn{1}{c|}{0/6}          & \multicolumn{1}{c|}{\textbf{75.61}} & 2/6          \\ \hline
MT (Simple) & \multicolumn{1}{c|}{86.87}          & \multicolumn{1}{c|}{0/6}          & \multicolumn{1}{c|}{69.07}          & 0/6          & \multicolumn{1}{c|}{87.53}          & \multicolumn{1}{c|}{0/6}          & \multicolumn{1}{c|}{73.26}          & 1/6          \\ \hline
MT (Ours)   & \multicolumn{1}{c|}{\textbf{87.98}} & \multicolumn{1}{c|}{\textbf{6/6}} & \multicolumn{1}{c|}{\textbf{72.86}} & \textbf{5/6} & \multicolumn{1}{c|}{\textbf{89.03}} & \multicolumn{1}{c|}{\textbf{6/6}} & \multicolumn{1}{c|}{74.44}          & \textbf{3/6} \\ \hline
\end{tabular}}
\label{table:simple_vs_complex}
\caption{A comparison of a single-task (ST), a simple multi-task (SMT), and our complex multi-task (CMT) model in default training and active learning setups. All models were trained using cross-entropy (CE) or label smoothing (LS).}
\end{table}
}

\com{
\begin{figure*}[!ht]
    \centering
\includegraphics[width=0.99\linewidth]{figures/simple_vs_complex_grid.png}
    \caption{ST, Simple MT, and Our MT.} 
    \label{fig:model_architecture}
\end{figure*}}

\section{MT-AL for Complementing Tasks}
\label{sec:closely_related_tasks}
We start by investigating MT-AL for two closely-related, complementing, syntactic tasks: Dependency Parsing (DP) and Named Entity Recognition (NER), which are often solved together by a joint multi-task model 
\cite{finkel2009joint, nguyen-nguyen-2021-phonlp}.

\subsection{Research Questions}
\label{sec:research_questions}

We focus on three research questions. At first, we would like to establish whether MT-AL methods are superior to ST-AL methods for multi-task learning. Our first two questions are hence:
\textbf{Q1.1}: Is multi-task learning effective in this setup? and \textbf{Q1.2}: Is AL effective? If so, which AL strategy is better: ST-AL or MT-AL?

Next, notice that in MT-AL the confidence score of an example can be based on one or more of the participating tasks. That is, even if the base model for which training examples are selected is an MTL model, the confidence scores used by the MT-AL algorithm can be based on one task or more (\S \ref{sec:mt-al-confidence}). Our third question is thus:
\textbf{Q1.3}: Is it better to calculate confidence scores based on one of the participating tasks, or should we consider a joint confidence score, based on both tasks?\footnote{This question naturally generalizes when more than two tasks are involved.}

\com{
Following our observation that cross-entropy (CE) training yields overconfident multi-task predictions ($\S$ \ref{sec:loss_funcs}), our next two questions consider the benefits of label smoothing (LS) for MT-AL, first for overconfidence reduction, and then for overall performance.
\textbf{Q1.4}: Can LS mitigate the overconfidence behavior of Transformer-based models in MT-AL setups? and
\textbf{Q1.5}: Does the in-domain performance improve when training with the LS objective instead of the CE objective?}


\subsection{Data}
We consider the English version of the OntoNotes 5.0 corpus \cite{hovy2006ontonotes}, consisting of seven textual domains: broadcast conversation (BC), broadcast news (BN), magazine (MZ), news (NW), bible (PT), telephone conversation (TC) and web (WB). Sentences are annotated with constituency-parse trees, named entities, part-of-speech tags, as well as other labels. We convert constituency-parse trees to dependency trees using the ElitCloud conversion tool.\footnote{\url{https://github.com/elitcloud/elit-java}.} 
We do not report results in the PT domain, as it is not annotated for NER. Table \ref{table:dataset_statistics} summarizes the number of sentences per split for the OntoNotes domains, as well as for the additional datasets used in our next setups.

\begin{table*}[!ht]
\scalebox{0.75}{
\begin{tabular}{|c|cccccc|ccccc|cc|}
\hline
      & \multicolumn{6}{c|}{DP-NER}                                                                                                                                 & \multicolumn{5}{c|}{NER-RE}                                                                                                   & \multicolumn{2}{c|}{SF-ID}          \\ \hline
      & \multicolumn{1}{c|}{BC}     & \multicolumn{1}{c|}{BN}     & \multicolumn{1}{c|}{MZ}    & \multicolumn{1}{c|}{NW}     & \multicolumn{1}{c|}{TC}     & WB     & \multicolumn{1}{c|}{NYT24}  & \multicolumn{1}{c|}{NYT29}  & \multicolumn{1}{c|}{ScieRC} & \multicolumn{1}{c|}{WebNLG} & WLP   & \multicolumn{1}{c|}{ATIS}  & SNIPS  \\ \hline
Train & \multicolumn{1}{c|}{11,877} & \multicolumn{1}{c|}{10,681} & \multicolumn{1}{c|}{6,771} & \multicolumn{1}{c|}{34,967} & \multicolumn{1}{c|}{12,889} & 15,639 & \multicolumn{1}{c|}{56,193} & \multicolumn{1}{c|}{63,305} & \multicolumn{1}{c|}{1,540}  & \multicolumn{1}{c|}{4,973}  & 6,690 & \multicolumn{1}{c|}{4,478} & 13,084 \\ \hline
Dev   & \multicolumn{1}{c|}{2,115}  & \multicolumn{1}{c|}{1,293}  & \multicolumn{1}{c|}{640}   & \multicolumn{1}{c|}{5,894}  & \multicolumn{1}{c|}{1,632}  & 2,264  & \multicolumn{1}{c|}{5,000}  & \multicolumn{1}{c|}{7,033}  & \multicolumn{1}{c|}{217}    & \multicolumn{1}{c|}{500}    & 2,320 & \multicolumn{1}{c|}{500}   & 700    \\ \hline
Test  & \multicolumn{1}{c|}{2,209}  & \multicolumn{1}{c|}{1,355}  & \multicolumn{1}{c|}{778}   & \multicolumn{1}{c|}{2,325}  & \multicolumn{1}{c|}{1,364}  & 1,683  & \multicolumn{1}{c|}{5,000}  & \multicolumn{1}{c|}{4,006}  & \multicolumn{1}{c|}{451}    & \multicolumn{1}{c|}{689}    & 2,343 & \multicolumn{1}{c|}{893}   & 700    \\ \hline
\end{tabular}}
\caption{Data statistics. We report the number of sentences in the original splits for each pair of tasks.}
\label{table:dataset_statistics}
\end{table*}

\com{
\begin{table}[!t]
   \scalebox{0.85}{
\begin{tabular}{|l|l|l|l|}
\hline
                                     & \textbf{Train} & \textbf{Dev} & \textbf{Test} \\ \hline
\textbf{Broadcast Conversation (BC)} & 173K           & 30K          & 36K           \\ \hline
\textbf{Broadcast News (BN)}         & 207K           & 25K          & 26K           \\ \hline
\textbf{Magazine (MZ)}               & 161K           & 15K          & 17K           \\ \hline
\textbf{News (NW)}                   & 878K           & 148K         & 60K           \\ \hline
\textbf{Telephone Conversation (TC)} & 92K            & 11K          & 11K           \\ \hline
\textbf{Web (WB)}                    & 361K           & 48K          & 50K           \\ \hline
\end{tabular}}
    \caption{OntoNotes statistics. We report the number of tokens per domain.}
    \label{table:dataset_statistics}
\end{table}
}

\com{
\begin{figure*}[!ht]
    \centering
    \subfloat[\centering A Single-task Model]
    {{\includegraphics[height=4cm,width=4cm]{figures/MTAL_Single_Architecture.png}}}
    \qquad
    \centering
    \subfloat[\centering A Multi-task Model]
    {{\includegraphics[height=4cm,width=8cm]{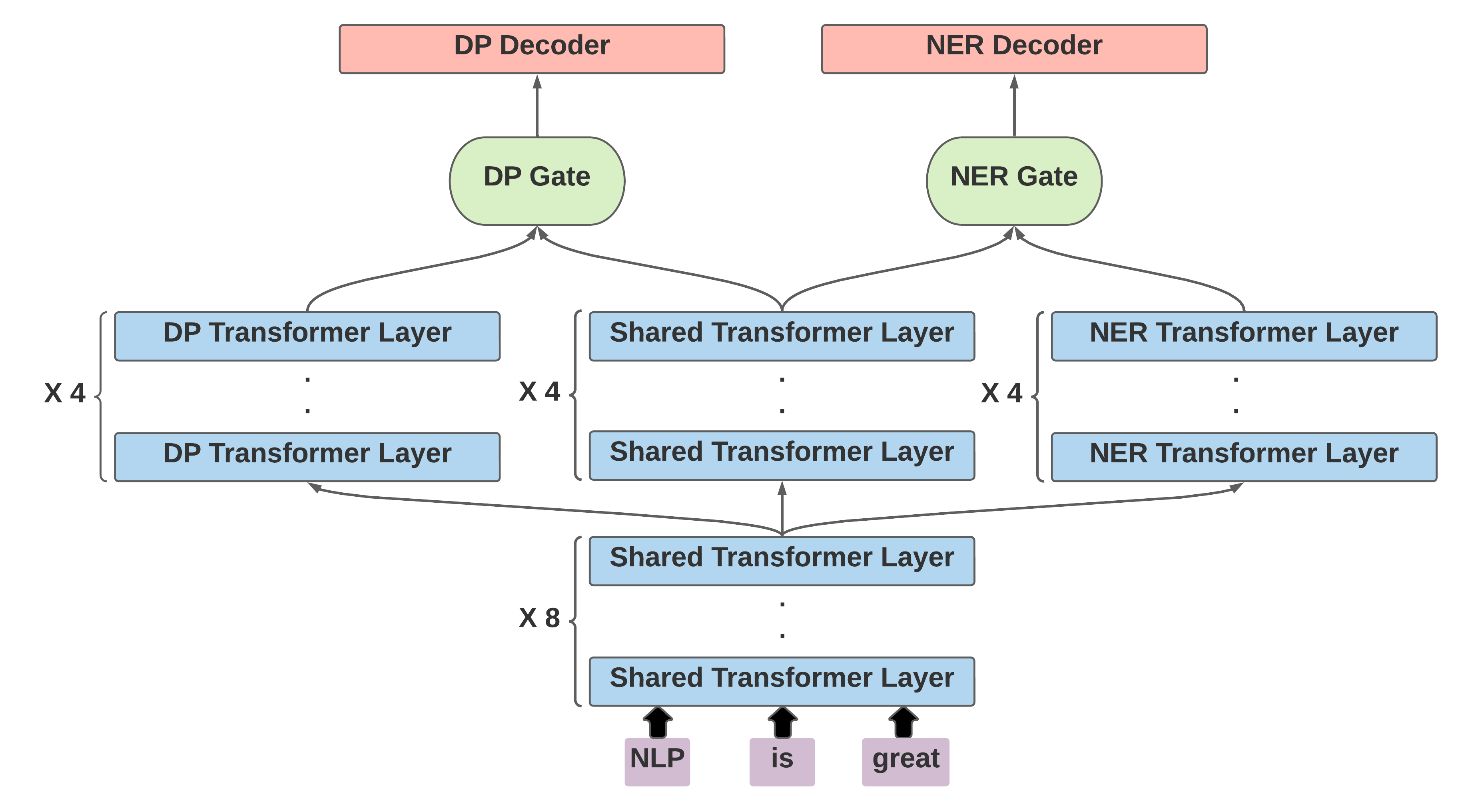}}}
    \caption{Our model architectures. (a) The single-task model is consisted of a 12-layer encoder and a single decoder; (b) The multi-task model consists of an encoder with 8 shared layers followed by 4 shared layers and 4 non-shared layers per task. The shared and non-shared representations are concatenated and passed through the gating mechanism before feeding the task-specific decoders.} 
    \label{fig:model_architecture}
\end{figure*}
}

\subsection{Models}
\label{sec:models}

We consider two model types: Single-task and multi-task models. Our single-task model (ST) consists of the 12-layer pre-trained BERT-base encoder \cite{devlin2019bert}, followed by a task decoder. At first, we implemented a simple multi-task model (SMT), consisting of a shared 12-layer pre-trained BERT-base encoder followed by an independent decoder for each task. However, early results suggested that it is inferior to single-task modeling. We therefore implemented a more complex multi-task model (CMT), illustrated in Figure \ref{fig:model_architecture}.
%
%
%
%
%
This model consists of (shared) cross-task and task-specific modules, similar in nature to the architecture proposed by \citet{lin2018multi}. 
In particular, it uses the 8 bottom BERT layers as shared cross-task layers and employs $t + 1$ replications of the 4 top BERT layers, one replication for each task, as well as a shared cross-task replication. The input text, as encoded by the shared 8 layers, $e_{1:8}^S$, is passed through the shared and non-shared 4-layer modules, $e_{8:12}^S$ and $e_{8:12}^{U^i}$, respectively. The task classifiers are then fed with the output of the cross-task layers combined with the output of their task-specific layers, following the gating mechanism of \citet{rotman2019deep}:
\begin{equation*}
 \begin{gathered}
 \scalebox{.92}{$a^i(x) = \sigma (W^i_g[e_{8:12}^{S}(x);e_{8:12}^{U^i}(x)] + b^i_g)$}, \\
 \scalebox{.92}{$g^i(x) = a^i(x) \odot e_{8:12}^{S}(x) + (1-a^i(x)) \odot e_{8:12}^{U^i}(x)$},
 \end{gathered}
\end{equation*}
where $;$ is the concatenation operator, $\odot$ is the element-wise product, $\sigma$ is the Sigmoid function, and $W^i_g$ and $b^i_g$ are the gating mechanism parameters. The combined vector $g^i(x)$ is then fed to the $i$-th task-specific decoder.\footnote{We considered several other parameter-sharing schemes but witnessed lower performance.}

All implementations are based on HuggingFace's Transformers package \cite{wolf2019huggingface}.\footnote{\url{https://github.com/huggingface/transformers}.} For all models, the DP decoder is based on the Biaffine parser \cite{dozat2017deep} and the NER decoder is a simple linear classifier.

\subsection{Training and Hyper-parameter Tuning}
\label{sec:exp-setup1}

We consider the following hyper-parameters for the AL experiments. 
At first, we randomly sample 2\% of the original training set to serve as the initial labeled examples in all experiments and treat the rest of the training examples as unlabeled. We also fix our development set to be twice the size of our initial training set, by randomly sampling examples from the original development set. We then run each AL method for 5 iterations, where at each iteration, the algorithm selects an unlabeled set of the size of its initial training set (that is, 2\% of the original training set) for annotation. We then reveal the labels of the selected examples and add them to the training set of the next iteration. At the beginning of the final iteration, our labeled training set consists of 10\% of the original training data.

In each iteration, we train the models with 20K gradient steps with an early stopping criterion according to the development set. We report LAS scores for DP and F1 scores for NER. For DP, we measure our AL confidence scores on the unlabeled edges. When performing multi-task learning, we set the stopping criterion as the geometric mean of the task scores (F1 for NER and LAS for DP). We optimize all parameters using the ADAM optimizer \cite{kingma2015adam} with a weight decay of 0.01, a learning rate of 5e-5, and a batch size of 32. For label smoothing (see below), we use $\alpha = 0.2$. \com{\footnote{We also experimented with $\alpha$-values of 0.1 and 0.3, but mostly witnessed lower results.}} Following \citet{dror2018hitchhiker}, we use the t-test for measuring statistical significance ($\textit{p-value}=0.05$).

\begin{figure}[!t]
    \centering
\includegraphics[width=\linewidth,height=6cm]{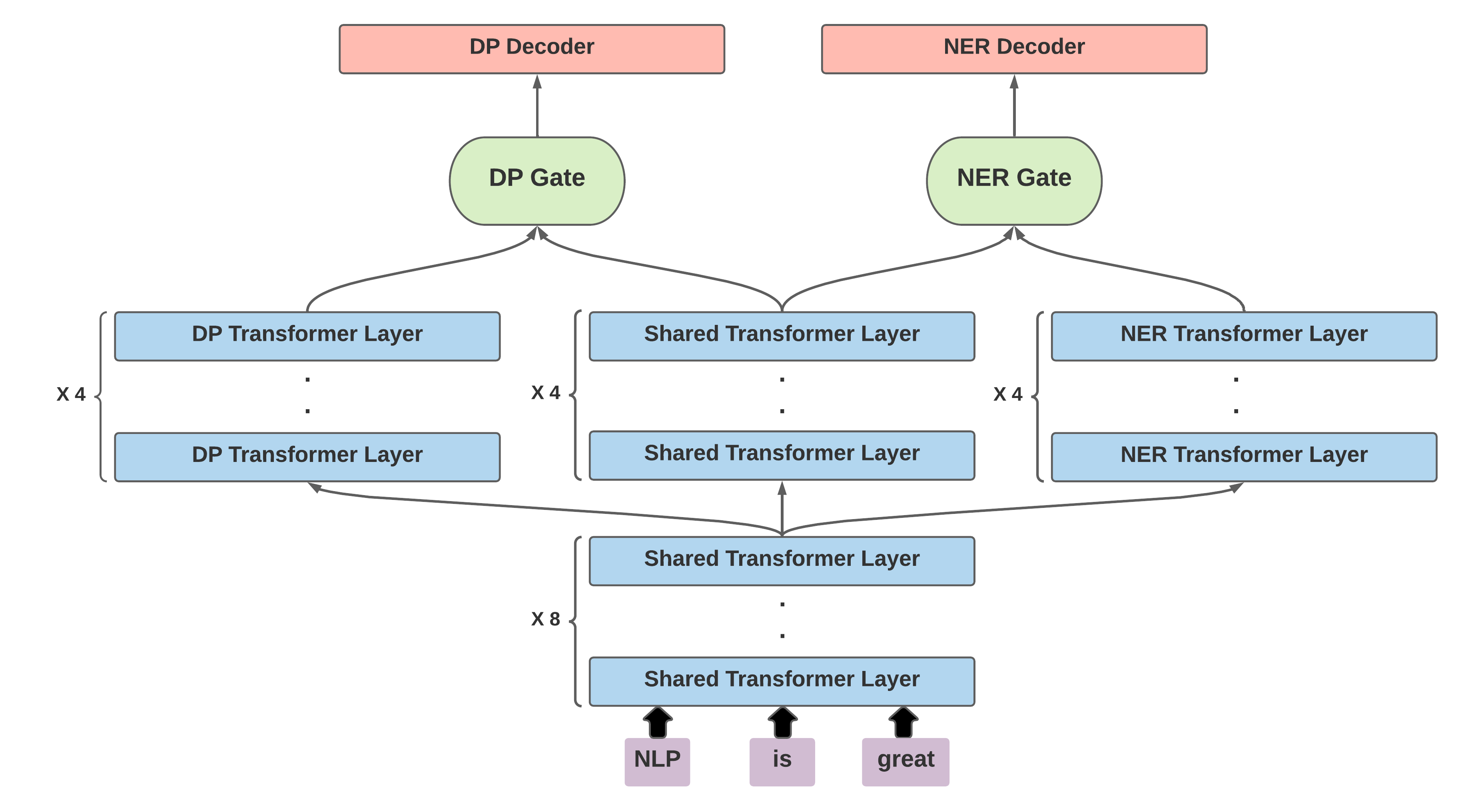}
    \caption{Our complex multi-task model architecture for DP and NER.} 
    \label{fig:model_architecture}
\end{figure}

\begin{table}[!t]
\centering
\begin{adjustbox}{width=\linewidth}
\begin{tabular}{|c|cccc|cccc|}
\hline
               & \multicolumn{4}{c|}{Full Training}                                                                                           & \multicolumn{4}{c|}{Active Learning}                                                                                         \\ \hline
               & \multicolumn{2}{c|}{DP}                                                 & \multicolumn{2}{c|}{NER}                           & \multicolumn{2}{c|}{DP}                                                 & \multicolumn{2}{c|}{NER}                           \\ \hline
               & \multicolumn{1}{c|}{Avg}            & \multicolumn{1}{c|}{Best}         & \multicolumn{1}{c|}{Avg}            & Best         & \multicolumn{1}{c|}{Avg}            & \multicolumn{1}{c|}{Best}         & \multicolumn{1}{c|}{Avg}            & Best         \\ \hline
ST (CE)        & \multicolumn{1}{c|}{\textbf{87.17}} & \multicolumn{1}{c|}{1/6}          & \multicolumn{1}{c|}{70.35}          & 0/6          & \multicolumn{1}{c|}{\textbf{86.43}} & \multicolumn{1}{c|}{\textbf{3/6}} & \multicolumn{1}{c|}{\textbf{74.65}} & 6/6          \\ \hline
SMT (CE) & \multicolumn{1}{c|}{86.94}          & \multicolumn{1}{c|}{2/6}          & \multicolumn{1}{c|}{67.51}          & 0/6          & \multicolumn{1}{c|}{85.86}          & \multicolumn{1}{c|}{2/6}          & \multicolumn{1}{c|}{70.31}          & 0/6          \\ \hline
CMT (CE)   & \multicolumn{1}{c|}{87.04}          & \multicolumn{1}{c|}{\textbf{3/6}} & \multicolumn{1}{c|}{\textbf{72.79}} & \textbf{6/6} & \multicolumn{1}{c|}{85.91}          & \multicolumn{1}{c|}{1/6}          & \multicolumn{1}{c|}{72.11}          & 0/6          \\ \hline \hline
ST (LS)        & \multicolumn{1}{c|}{87.64}          & \multicolumn{1}{c|}{0/6}          & \multicolumn{1}{c|}{71.31}          & 1/6          & \multicolumn{1}{c|}{88.96}          & \multicolumn{1}{c|}{0/6}          & \multicolumn{1}{c|}{\textbf{75.61}} & 2/6          \\ \hline
SMT (LS) & \multicolumn{1}{c|}{86.87}          & \multicolumn{1}{c|}{0/6}          & \multicolumn{1}{c|}{69.07}          & 0/6          & \multicolumn{1}{c|}{87.53}          & \multicolumn{1}{c|}{0/6}          & \multicolumn{1}{c|}{73.26}          & 1/6          \\ \hline
CMT (LS)   & \multicolumn{1}{c|}{\textbf{87.98}} & \multicolumn{1}{c|}{\textbf{6/6}} & \multicolumn{1}{c|}{\textbf{72.86}} & \textbf{5/6} & \multicolumn{1}{c|}{\textbf{89.03}} & \multicolumn{1}{c|}{\textbf{6/6}} & \multicolumn{1}{c|}{74.44}          & \textbf{3/6} \\ \hline
\end{tabular}
\end{adjustbox}
\caption{A comparison of a single-task model (ST), a simple multi-task model (SMT), and our complex multi-task model (CMT) in full training and active learning. Models were trained with the cross-entropy (CE) or label smoothing (LS) losses, on all OntoNotes domains.}
\label{table:simple_vs_complex}
\end{table}

\subsection{Results}
\label{sec:results}

\com{
\begin{figure}[!ht]
    \centering
    \subfloat[\centering Dependency Parsing]
    {{\includegraphics[width=0.5\textwidth]{figures/dp_main_results.png}}}
    \qquad
    \centering
    \subfloat[\centering Named Entity Recognition]
    {{\includegraphics[width=0.5\textwidth]{figures/ner_main_results.png}}}
    \caption{Main Results.} 
    \label{fig:main_results}
\end{figure}
}

\paragraph{Model Architecture (Q1.1)}
We would first like to investigate the performance of the single-task and multi-task models in the full training (FT) and, more importantly, in the active learning (AL) setups. We hence compare three architectures: The single-task model (ST), the simple multi-task model (SMT), and our complex multi-task model (CMT). We train each model for DP and for NER on the six OntoNotes domains using the cross-entropy (CE) objective function, or with the label smoothing objective (LS \cite{szegedy2016rethinking}) that has been demonstrated to decrease calibration errors of Transformer models \cite{desai2020calibration, kong-etal-2020-calibrated}. ST-AL is performed with \textit{ST-EC} and MT-AL with \textit{MT-AVG}.

Table \ref{table:simple_vs_complex} reports the average scores (Avg column) over all domains and the number of domains where each model achieved the best results (Best). The results raise three important observations. First, SMT is worse on average than ST in all setups, suggesting that \textbf{\textit{vanilla MT is not always better than ST training}}. Second, \textbf{\textit{our CMT model achieves the best scores in most cases}}. The only case where it is inferior to ST, but not to SMT, is on AL with CE training. However, when training with LS, it achieves results comparable to or higher than those of ST on AL.

Third, when comparing CE to LS training, \textbf{\textit{LS clearly improves the average scores of all models}} (besides one case). Interestingly, the improvement is more significant in the AL setup than in the FT setup. We report that when expanding these experiments to all AL selection methods, LS was found very effective for both tasks, outperforming CE in most comparisons, with an average improvement of 1.8\% LAS for DP and 0.9 F1 for NER.


\paragraph{Multi-task vs. Single-task Performance (Q1.2)}

\begin{figure}[!t]
    \centering
    \subfloat[\centering Dependency Parsing]
    {{\includegraphics[width=\linewidth,height=4.0cm]{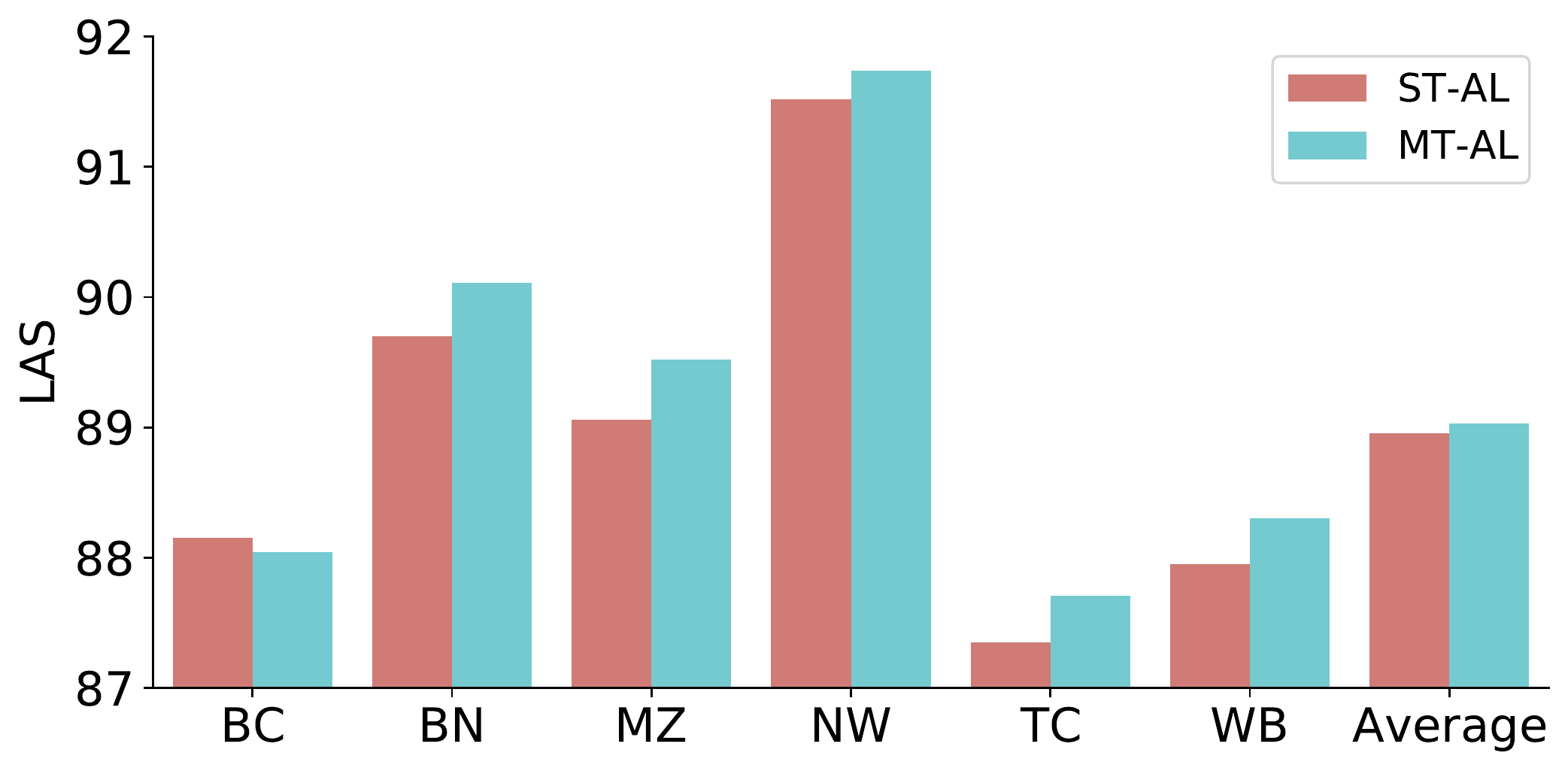}}}
    \qquad
    \centering
    \subfloat[\centering Named Entity Recognition]
    {{\includegraphics[width=\linewidth,height=4.0cm]{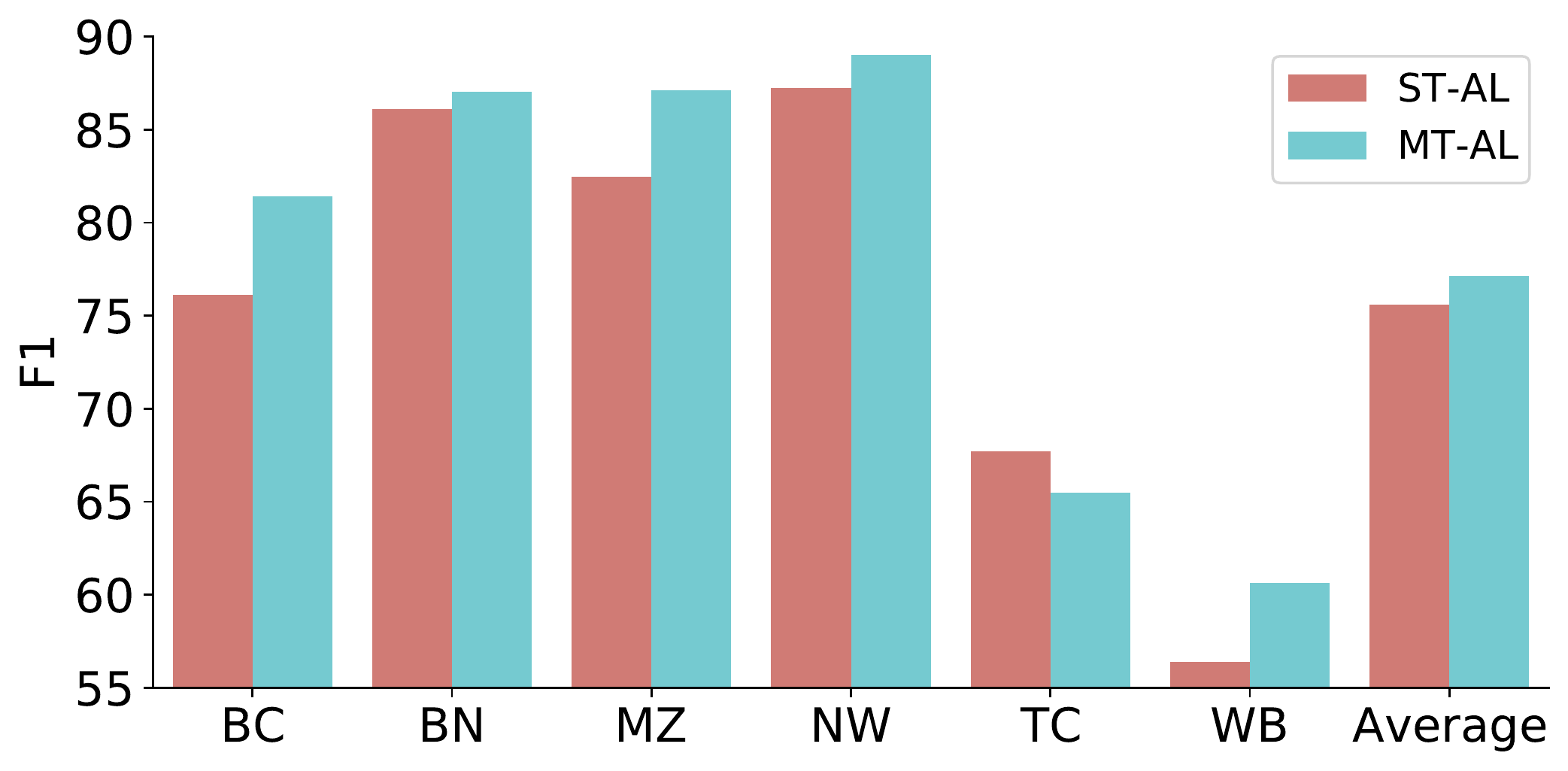}}}
    \caption{Performance of the best ST-AL vs. the best MT-AL method per domain (Q1.2).} 
    \label{fig:st_vs_mt_in_domain}
\end{figure}

We next ask whether MT-AL outperforms strong ST-AL baselines. Figure \ref{fig:st_vs_mt_in_domain} presents for every task and domain the performance of the per-domain best ST-AL and MT-AL methods after the final AL iteration. Following our observations in Q1.1, we train all models with the LS objective and base the multi-task models on the effective CMT model.

Although there is no single method, MT-AL or ST-AL, which performs best across all domains and tasks, MT-AL seems to perform consistently better. The figure suggests that \textbf{\textit{MT-AL is effective for both tasks, outperforming the best ST-AL methods in 4 of 6 DP domains}} (results are not statistically significant, the average \textit{p-value} is 0.19) \textbf{\textit{and in 5 of 6 NER domains}} (results for 3 domains are statistically significant). While the average gap between MT-AL and ST-AL is small for DP (0.28\% LAS), in NER it is as high as 2.4 F1 points in favor of MT-AL. In fact, for half of the NER domains, this gap is greater than 4.2 F1.

When comparing individual selection methods, \textit{MT-AVG}, and multi-task DP-based entropy, \textit{MT-EC-DP}, are the best selection methods for DP, with average scores of 89.03\% and 88.99\%, respectively. Single-task DP-based entropy, \textit{ST-EC-DP} is third, with an average score of 88.96\%, while the second best ST-AL method, \textit{ST-DA-DP}, is ranked only ninth among all methods, outperformed by seven different MT-AL methods. For NER, multi-task NER-based entropy, \textit{MT-EC-NER}, is the best model with an average F1 score of 77.13, followed by \textit{MT-MAX} with an average F1 score of 75.90. The single-task NER-based methods, \textit{ST-EC-NER} and \textit{ST-DA-NER} are ranked only fifth and sixth both with an average score of 75.60. These results provide an additional indication of the superiority of MT-AL.

\guyfix{In terms of the MT-AL selection methods that do not perform a simple aggregation, \textit{MT-PAR} and \textit{MT-RRF} perform similarly, averaging 88.31\% LAS for DP and 75.88 F1 for NER, while \textit{MT-IND} achieves poor results for DP and moderate results for NER (an overall comparison of the MT-AL methods is provided in $\S$ 8)).}

\begin{figure}
\centering
\includegraphics[width=0.99\linewidth]{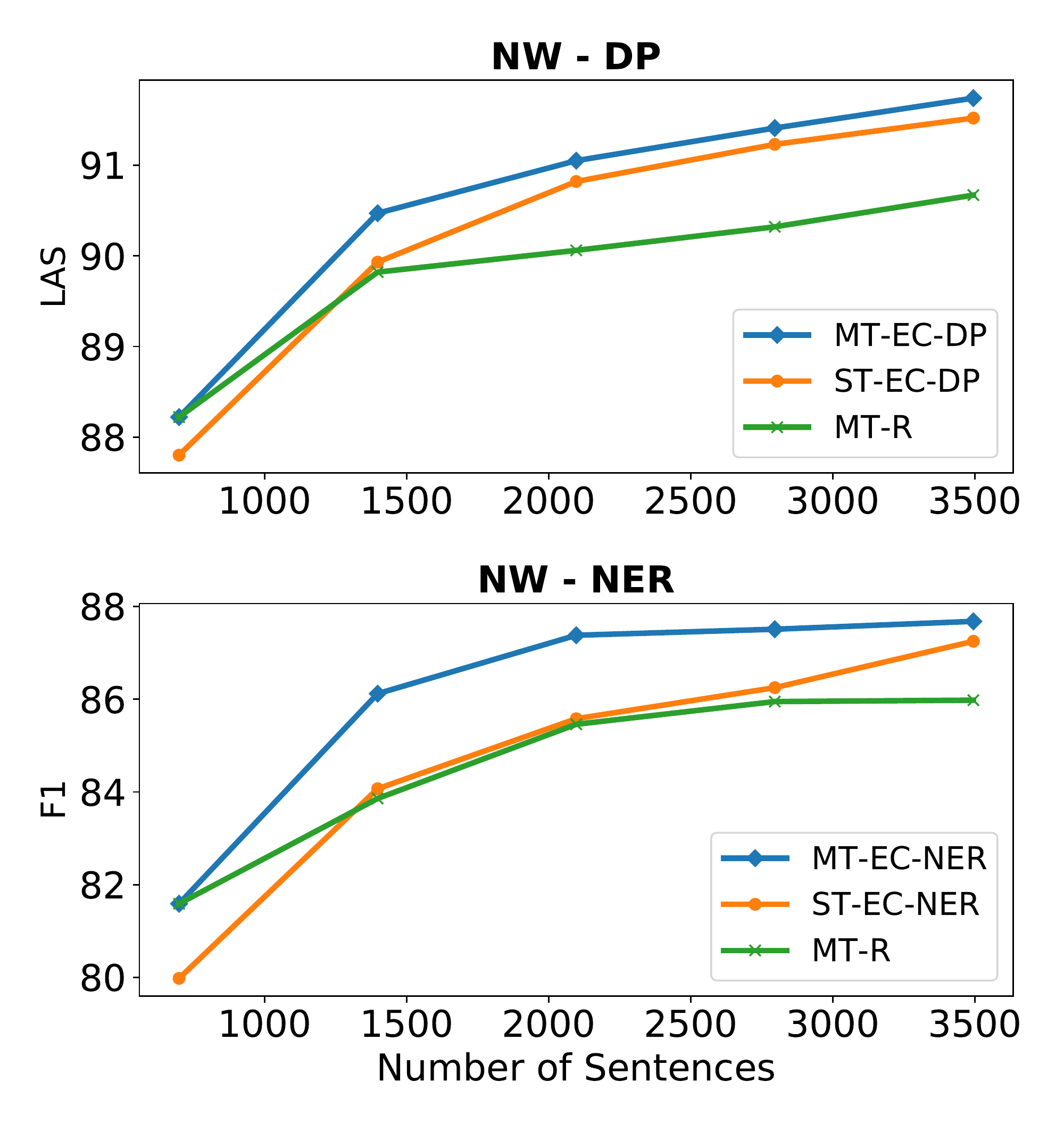}
\caption{Performance as a function of the number of training examples (Q1.2).}
\label{fig:main_results_entropy_plot}
\end{figure}

\com{
\begin{figure}[!t]
    \centering
    \subfloat[\centering Dependency Parsing]
    {{\includegraphics[width=0.5\textwidth,height=5.0cm]{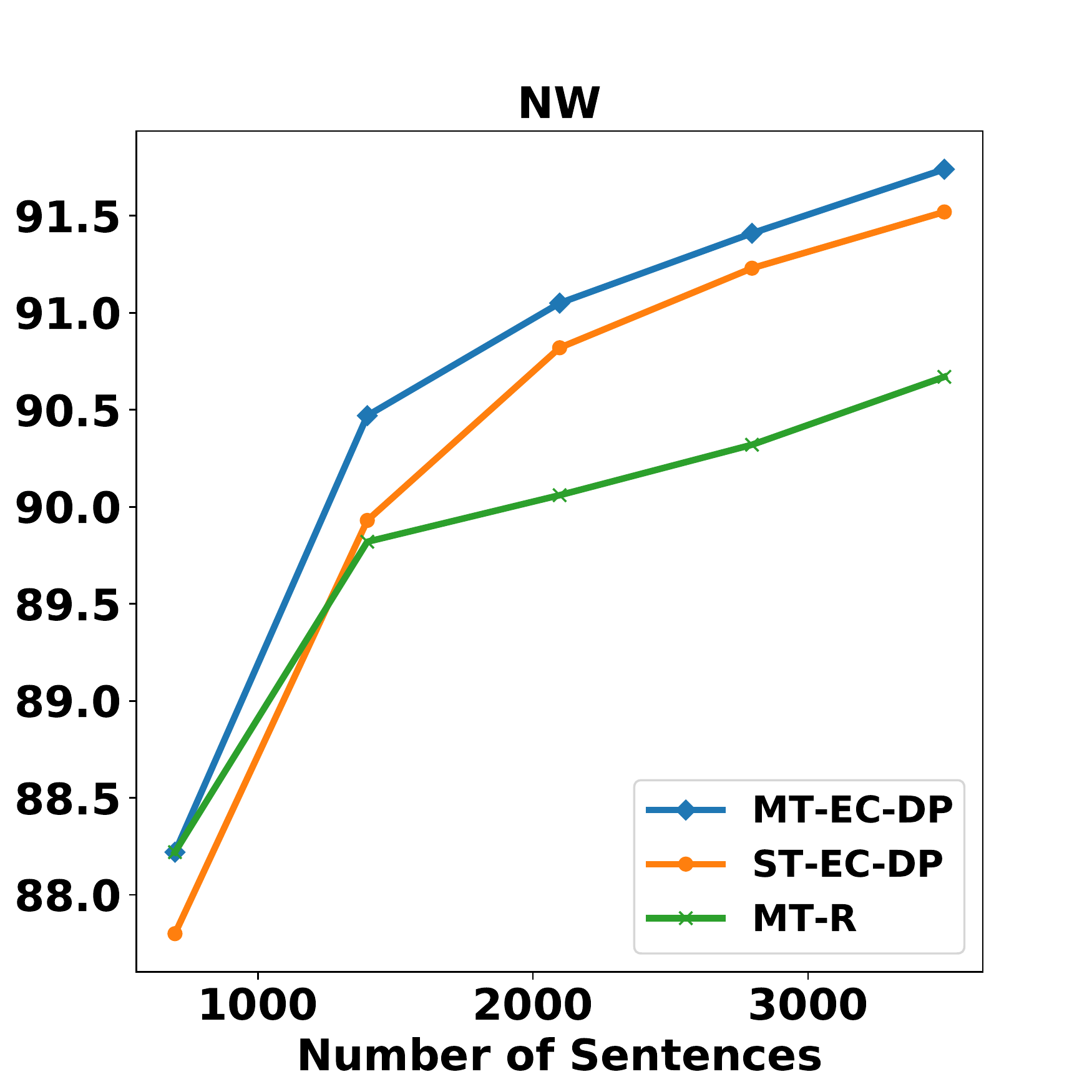}}}
    \qquad
    \centering
    \subfloat[\centering Named Entity Recognition]
    {{\includegraphics[width=0.5\textwidth,height=5.0cm]{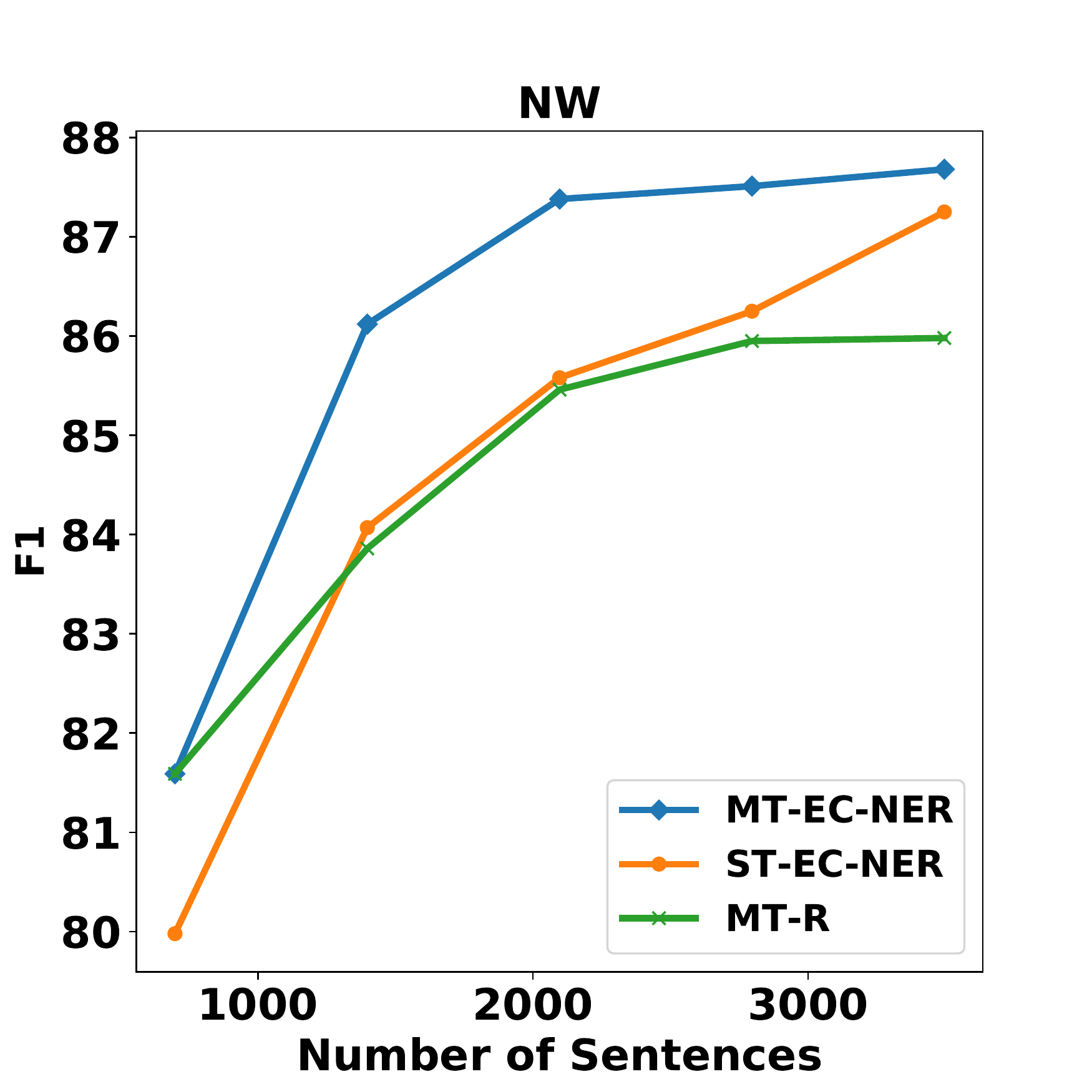}}}
    \caption{Performance as a function of the number of training examples (Q1.2).} 
    \label{fig:main_results_entropy_plot}
\end{figure}
}

\begin{table}[!t]
\begin{adjustbox}{width=\linewidth}
\begin{tabular}{|c|c|c|c|c|c|}
\hline
                     & \multicolumn{2}{c|}{\textbf{Within-task}} & \multicolumn{2}{c|}{\textbf{Cross-task}} & \textbf{Average}  \\ \hline
                     & \textbf{DP}         & \textbf{NER}        & \textbf{DP}        & \textbf{NER}        & \textbf{DP + NER} \\ \hline
MT-AL winning \% & 47.22               & 63.88               & 90.74              & 89.81              & 78.78            \\ \hline
\end{tabular}
\end{adjustbox}
\caption{A comparison of MT-AL vs. ST-AL on within-task, cross-task and average performance. Values indicate the percentage of comparisons in which MT-AL methods were superior.}
\label{table:st_vs_mt_in_domain_comparisons}
\end{table}

We next compare the per-iteration performance of ST-AL and MT-AL. To this end, Figure \ref{fig:main_results_entropy_plot} presents the performance for the most prominent MT-AL and ST-AL methods: \textit{MT-EC-DP} and \textit{ST-EC-DP} for DP and \textit{MT-EC-NER} and \textit{ST-EC-NER} for NER, together with the multi-task random selection method \textit{MT-R}. We plot for each method its task score on the NW domain (the one with the largest dataset) as a function of the training set size, corresponding to 2\% to 10\% of the original training examples. \textbf{\textit{Clearly, the MT-AL methods are superior across all AL iterations, indicating the stability of MT-AL as well as its effectiveness in low-resource setups}}. Similar patterns are also observed in the other domains.

As a final evaluation for Q1.2, we directly compare pairs of MT-AL and ST-AL methods, performing three comparison types on each of the domains: \textbf{Within-task}: Comparing the performance of \textit{ST-EC-i} and \textit{ST-DA-i} to their MT-AL counterparts (\textit{MT-EC-i} and \textit{MT-DA-i}) and to the joint-selection methods on task \textit{i}, either DP or NER (108 comparisons); \textbf{Cross-task}: Comparing the performance of \textit{ST-EC-i} and \textit{ST-DA-i} to their MT-AL counterparts and to the joint-selection methods on the opposite task (e.g., if the models select according to DP, we evaluate the NER task; 108 comparisons). This comparison allows us to evaluate the effect of single- and multi- task modeling on cross-task performance. Since single-task models cannot be directly applied to the opposite task, we record the examples selected by the ST-AL method and train a model for the opposite task on these examples; and \textbf{Average}: Comparing all ST-AL methods to all MT-AL methods according to their average performance on both tasks (264 comparisons).

Table \ref{table:st_vs_mt_in_domain_comparisons} reports the percentage of comparisons where the MT-AL methods are superior. On average, the two method types are on par when comparing \textbf{Within-task} performance. More interestingly, for \textbf{Cross-task} performance MT-AL methods are clearly superior with around 90\% winnings (87\% of the cases statistically significant). Finally, the \textbf{Average} also supports the superiority of MT-AL methods which perform better in 79\% of the cases (all results are statistically significant). \textbf{\textit{These results demonstrate the superiority of MT-AL, particularly (and perhaps unsurprisingly) when both tasks are considered.}}

\com{We hence provide a positive answer to Q1.2 and conclude that MT-AL is indeed better, especially when considering the performance of all tasks.}

\com{
We further continue to compare ST-AL to MT-AL by exploring performance over time. To this end, we present in Figure \ref{fig:main_results_entropy_plot} the two best ST-AL and MT-AL methods per task along with the random selection methods for four selected domains. We plot for each method its task score as a function of the training set size, corresponding to 2\% to 10\% of the original training samples.

For DP, the plots confirm that the two MT-AL methods, \textit{MT-AVG} and \textit{MT-EC-DP}, are the best models overall, surpassing the ST-AL and the random selection methods on all presented domains. Interestingly, the gap between the MT-AL methods and their ST-AL competitors is more substantial in the first three iterations, suggesting that MT-AL is highly effective for low-resource scenarios.

We witness similar patterns in NER. \textit{MT-EC-NER} and \textit{MT-MAX} outperform single-task and random selection methods in all presented domains and in all iterations. The only exceptions are the third and fourth iterations of the BN domain, where \textit{ST-DA-NER} surpasses the \textit{MT-EC-NER} method but is still outperformed by the \textit{MT-MAX} method. We again see the importance of the MT-AL methods in the first few iterations, where the gap from the other methods is clear. 

Not surprisingly, the two random selection methods \textit{ST-R} and \textit{MT-R} are the worst performing baselines. They achieve similar results on DP, however \textit{MT-R} is clearly better than \textit{ST-R} for NER, hence suggesting that NER benefits from the parsing representations of the multi-task models, while the named-entity representations are less effective for parsing performance.

We, therefore, grant a positive answer to our first question and conclude that MT-AL methods are indeed effective for AL applications in NLP, surpassing strong ST-AL methods especially when training data is scarce.}

\paragraph{Single-task vs. Joint-task Selection (Q1.3)}

Next, we turn to our third question which compares single-task vs. joint-task confidence scores. That is, we ask whether MT-AL methods that base their selection criterion on more than one task are better than ST-AL and MT-AL methods that compute confidence scores using a single task only. 

\begin{table}[!t]
\centering
\begin{tabular}{|c|c|c|c|}
\hline
\multicolumn{1}{|l|}{} & DP             & NER            & Average        \\ \hline
\textit{ST-EC-DP}               & 88.96         & 71.34          & 80.15          \\ \hline
\textit{ST-EC-NER}              & 86.66          & 75.75          & 81.21          \\ \hline
\textit{MT-EC-DP}               & 88.99          & 73.75          & 81.37          \\ \hline
\textit{MT-EC-NER}              & 86.90          & \textbf{77.13} & 82.01          \\ \hline
\textit{MT-AVG}                 & \textbf{89.02} & 74.44          & 81.74          \\ \hline
\textit{MT-MAX}                & 88.64          & 75.90          & \textbf{82.27} \\ \hline
\end{tabular}
\caption{A comparison of AL methods that base their selection on a single-task vs. joint-task confidence scores. Results are averaged across the OntoNotes domains (Q1.3). \com{The \textit{Average} column indicate average values over both tasks (Q1.3).}}
\label{table:single_selection_vs_joint_selection}
\end{table}

To answer this question, we compare the two best ST-AL and MT-AL methods that are based on single-task selection to the two best joint-task selection MT-AL methods. As previously, all methods employ the LS objective. Table \ref{table:single_selection_vs_joint_selection} reports the average scores (across domains) of each of these methods for DP, NER, and the average task score, based on the final AL iteration. 

While the method that performs best on average on both tasks is \textit{MT-MAX}, a joint-selection method, the second best method is \textit{MT-EC-NER}, a single-task selection method, and the gap is only 0.26 points. Not surprisingly, performance is higher when the evaluated task also serves as the task that the confidence score is based on, either solely or jointly with another task. \com{As such, the joint-selection method \textit{MT-AVG} is the best average performing method for DP, and the single-task selection method \textit{MT-EC-NER} is the best average performing method for NER.}

\textbf{\textit{Although the joint-selection methods are effective for both tasks, we cannot decisively conclude that they are better than MT-AL methods that perform single-task selection.}} However, we do witness another confirmation for our answer to Q1.2, as all presented MT-AL methods perform better on average on both tasks (the \textit{Average} column) than the ST-AL methods. 

\com{This is because ST-AL methods suffer from degradation in cross-task performance. As evidence, \textit{MT-AVG} and \textit{MT-MAX} surpass ST-AL methods in cross-task performance in all domains, except for a single case (results are statistically significant throughout).}

\com{This is since when evaluated on the second task, ST-AL methods perform poorly and are not able to generalize beyond the task they were originally trained on.}

\com{
\begin{table}[!t]
\scalebox{0.55}{
\begin{tabular}{cccccccc}
\hline
\multicolumn{8}{|c|}{\textbf{Dependency Parsing}} \\ \hline
\multicolumn{1}{|c|}{Model / Domain} &
  \multicolumn{1}{c|}{BC} &
  \multicolumn{1}{c|}{BN} &
  \multicolumn{1}{c|}{MZ} &
  \multicolumn{1}{c|}{NW} &
  \multicolumn{1}{c|}{TC} &
  \multicolumn{1}{c|}{WB} &
  \multicolumn{1}{c|}{Average} \\ \hline
\multicolumn{1}{|c|}{ST-EC-DP (CE)} &
  \multicolumn{1}{c|}{0.0720} &
  \multicolumn{1}{c|}{0.0672} &
  \multicolumn{1}{c|}{0.0757} &
  \multicolumn{1}{c|}{0.0506} &
  \multicolumn{1}{c|}{0.0584} &
  \multicolumn{1}{c|}{0.3064} &
  \multicolumn{1}{c|}{0.1050} \\ \hline
\multicolumn{1}{|c|}{ST-EC-DP (LS)} &
  \multicolumn{1}{c|}{\textbf{0.0269}} &
  \multicolumn{1}{c|}{\textbf{0.0186}} &
  \multicolumn{1}{c|}{\textbf{0.0122}} &
  \multicolumn{1}{c|}{\textbf{0.0107}} &
  \multicolumn{1}{c|}{\textbf{0.0289}} &
  \multicolumn{1}{c|}{\textbf{0.1725}} &
  \multicolumn{1}{c|}{\textbf{0.0450}} \\ \hline
 &
   &
   &
   &
   &
   &
   &
   \\ \hline
\multicolumn{8}{|c|}{\textbf{Named Entity Recognition}} \\ \hline
\multicolumn{1}{|c|}{Model / Domain} &
  \multicolumn{1}{c|}{BC} &
  \multicolumn{1}{c|}{BN} &
  \multicolumn{1}{c|}{MZ} &
  \multicolumn{1}{c|}{NW} &
  \multicolumn{1}{c|}{TC} &
  \multicolumn{1}{c|}{WB} &
  \multicolumn{1}{c|}{Average} \\ \hline
\multicolumn{1}{|c|}{ST-EC-NER (CE)} &
  \multicolumn{1}{c|}{0.0111} &
  \multicolumn{1}{c|}{0.0167} &
  \multicolumn{1}{c|}{0.0131} &
  \multicolumn{1}{c|}{0.0146} &
  \multicolumn{1}{c|}{0.0090} &
  \multicolumn{1}{c|}{0.0132} &
  \multicolumn{1}{c|}{0.0129} \\ \hline
\multicolumn{1}{|c|}{ST-EC-NER (LS)} &
  \multicolumn{1}{c|}{\textbf{0.0002}} &
  \multicolumn{1}{c|}{\textbf{0.0002}} &
  \multicolumn{1}{c|}{\textbf{0.0001}} &
  \multicolumn{1}{c|}{\textbf{0.0002}} &
  \multicolumn{1}{c|}{\textbf{0.0009}} &
  \multicolumn{1}{c|}{\textbf{0.0010}} &
  \multicolumn{1}{c|}{\textbf{0.0004}} \\ \hline
 &
   &
   &
   &
   &
   &
   &
   \\ \hline
\multicolumn{8}{|c|}{\textbf{Dependency Parsing + Named Entity Recognition}} \\ \hline
\multicolumn{1}{|l|}{Model / Domain} &
  \multicolumn{1}{c|}{BC} &
  \multicolumn{1}{c|}{BN} &
  \multicolumn{1}{c|}{MZ} &
  \multicolumn{1}{c|}{NW} &
  \multicolumn{1}{c|}{TC} &
  \multicolumn{1}{c|}{WB} &
  \multicolumn{1}{c|}{Average} \\ \hline
\multicolumn{1}{|c|}{MT-AVG (CE)} &
  \multicolumn{1}{c|}{0.0436} &
  \multicolumn{1}{c|}{0.0499} &
  \multicolumn{1}{c|}{0.0473} &
  \multicolumn{1}{c|}{0.0345} &
  \multicolumn{1}{c|}{0.0393} &
  \multicolumn{1}{c|}{0.1681} &
  \multicolumn{1}{c|}{0.0638} \\ \hline
\multicolumn{1}{|c|}{MT-AVG (LS)} &
  \multicolumn{1}{c|}{\textbf{0.0040}} &
  \multicolumn{1}{c|}{\textbf{0.0018}} &
  \multicolumn{1}{c|}{\textbf{0.0013}} &
  \multicolumn{1}{c|}{\textbf{0.0012}} &
  \multicolumn{1}{c|}{\textbf{0.0057}} &
  \multicolumn{1}{c|}{\textbf{0.0546}} &
  \multicolumn{1}{c|}{\textbf{0.0114}} \\ \hline
\end{tabular}
}
\caption{Overconfidence results.}
\label{table:overconfidence}
\end{table}
}

\begin{figure}[!t]
    \centering
    \subfloat[\centering DP]
    {{\includegraphics[width=0.45\linewidth,height=3.5cm]{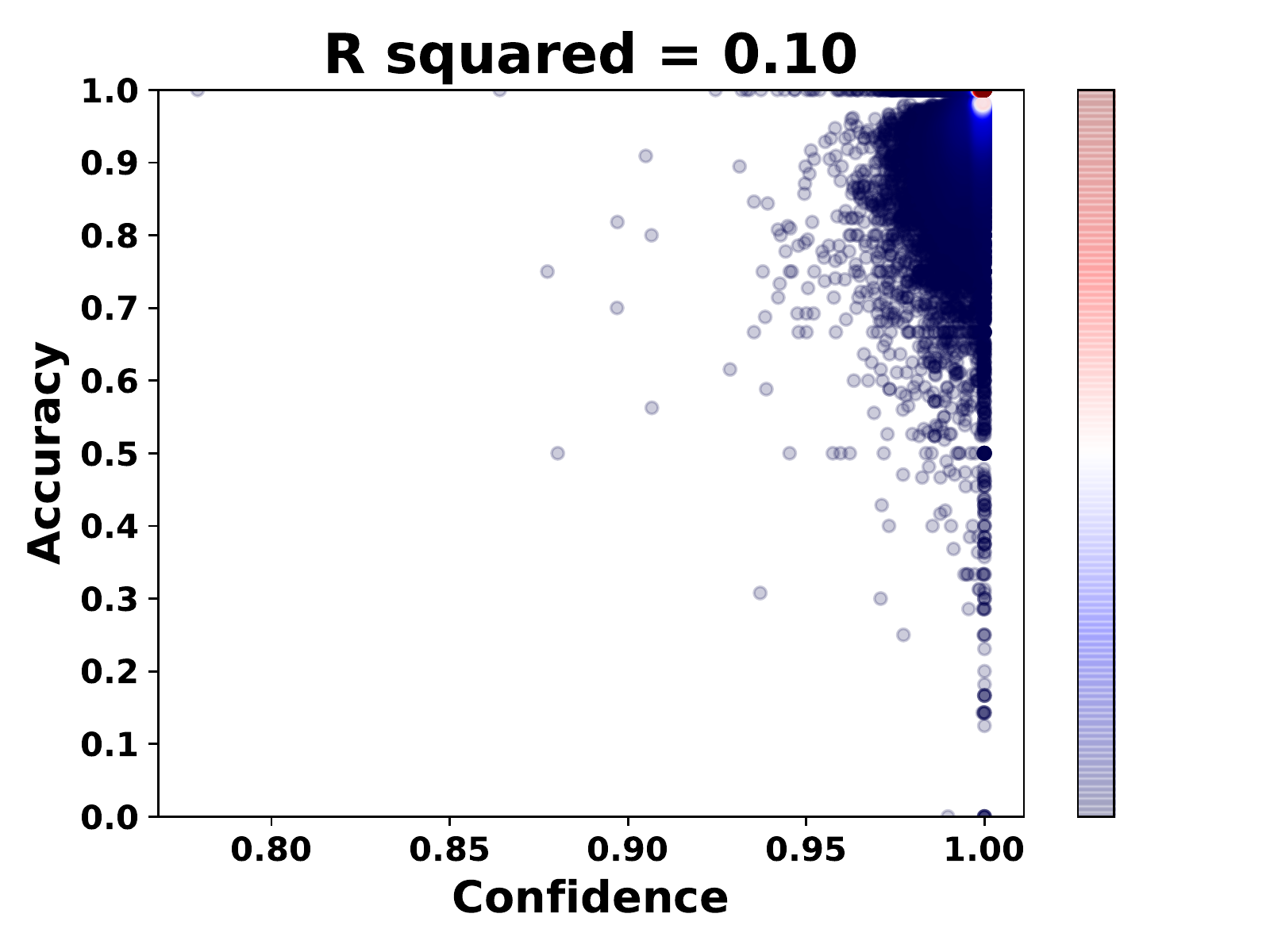}}}
    \qquad
    \centering
    \subfloat[\centering NER]
    {{\includegraphics[width=0.45\linewidth,height=3.5cm]{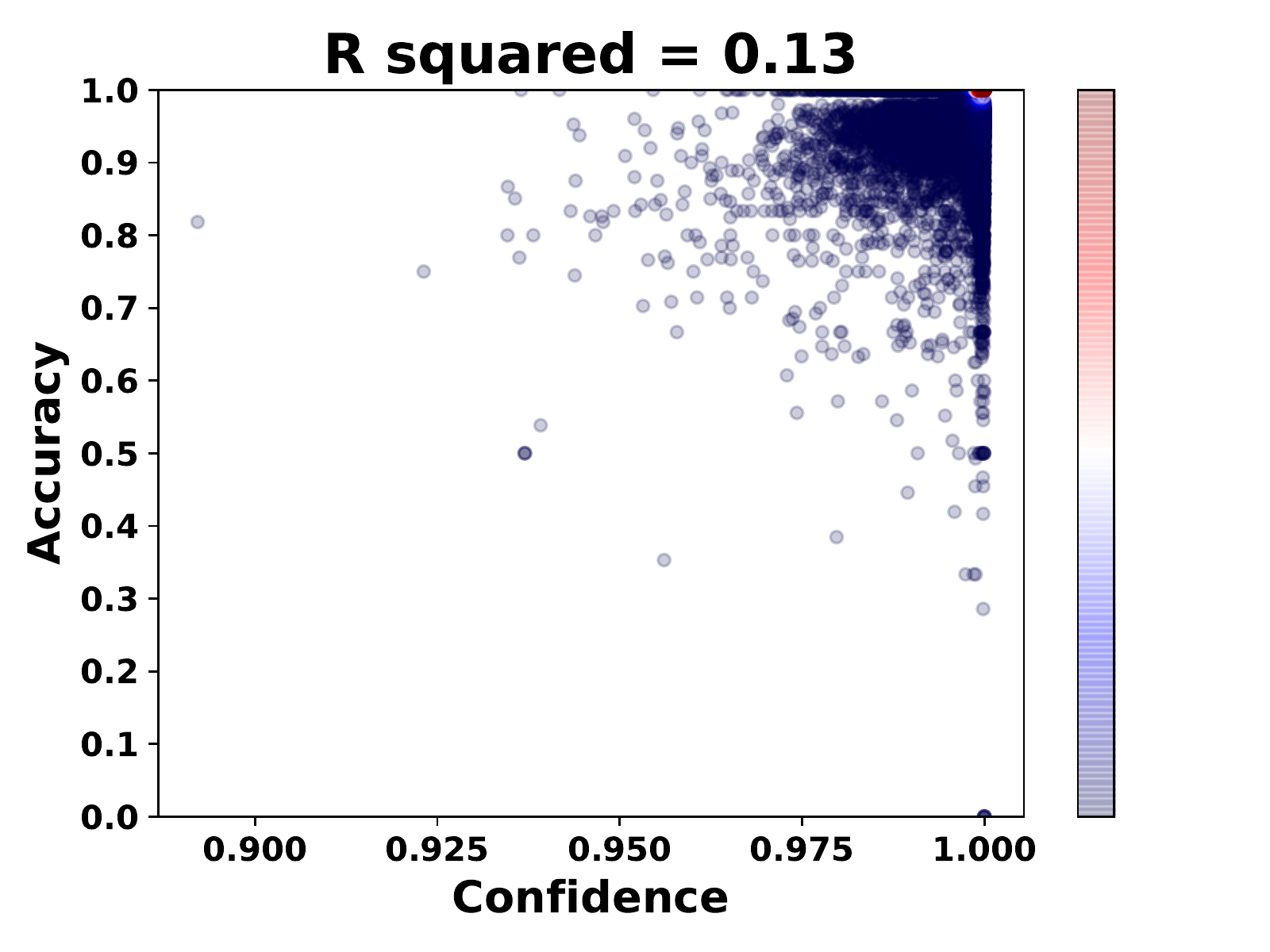}}}
    \caption{Sentence-level accuracy as a function of entropy-based confidence, for DP (left) and for NER (right), when training with the CE objective. The heat maps represent the point frequency.} 
    \label{fig:acc_vs_conf_ce}
\end{figure}

\paragraph{Overconfidence Analysis}

Originally, we trained our models with the standard CE loss. However, our early experiments suggested that such CE-based training yields overconfident models, which is likely to severely harm confidence-based AL methods. While previous work demonstrated the positive impact of label smoothing (LS) on model calibration, to the best of our knowledge, the resulting impact on multi-task learning has not been explored, specifically not in the context of AL. We next analyze this impact, which is noticeable in our above results, in more detail.

Figure \ref{fig:acc_vs_conf_ce} presents sentence-level confidence scores as a function of sentence-level accuracy when separately training a single-task BERT-base model on DP (left figure) and on NER (right figure) with the CE objective. The confidence scores were computed according to the $\textit{ST-EC}$ scores. The figure confirms that the model tends to be overconfident in its predictions. Furthermore, the low $R^2$ values (0.1 and 0.13 for DP and NER, respectively) indicate poor model calibration, since confidence scores are not correlated with actual accuracy. Similar patterns were observed when training our multi-task models with the CE objective.

Following this analysis, we turn to investigate the impact of LS on model predictions in MT-AL. Inspired by \citet{sunilmixup2019}, who defined the \textit{overconfidence error} ($OE$) for classification tasks, we first slightly generalize $OE$ to support sentence-level scores for token classification tasks. Given $N$ sentences, for each sentence $x$ we start by calculating its accuracy score $acc(x)$ over its tokens. 
The confidence score $conf(x)$ is set to the confidence score of the corresponding AL method. We then define $OE$ as:
\begin{equation*}
\resizebox{0.99\hsize}{!}{
    $OE = \frac{1}{N} \sum_{x=1}^{N} conf(x) \times max \big( conf(x) - acc(x), 0 \big)$.
    }
\end{equation*}
In essence, $OE$ penalizes predictions according to the gap between their confidence score and their accuracy, but only when the former is higher.

\begin{table}[!t]
\centering
\begin{adjustbox}{width=\linewidth}
\begin{tabular}{|c|c|c|c|c|c|c|c|}
\hline
\multicolumn{8}{|c|}{\textbf{DP}}                                                                                                              \\ \hline
                 & BC              & BN              & MZ              & NW              & TC              & WB              & Average         \\ \hline
\textit{ST-EC-DP} (CE)  & 0.0720          & 0.0672          & 0.0757          & 0.0506          & 0.0584          & 0.3064          & 0.1051          \\ \hline
\textit{ST-EC-DP} (TS)  & 0.0753          & 0.0651          & 0.0673          & 0.0421          & 0.0545          & 0.3012          & 0.1009          \\ \hline
\textit{ST-EC-DP} (LS)  & \textbf{0.0269} & \textbf{0.0186} & \textbf{0.0122} & \textbf{0.0107} & \textbf{0.0289} & 0.1725          & \textbf{0.0450} \\ \hline
\textit{ST-DA-DP} (CE)  & 0.0410          & 0.0392          & 0.0414          & 0.0337          & 0.0381          & \textbf{0.1098} & 0.0505          \\ \hline
\multicolumn{8}{|c|}{}                                                                                                                         \\ \hline
\multicolumn{8}{|c|}{\textbf{NER}}                                                                                                             \\ \hline
                 & BC              & BN              & MZ              & NW              & TC              & WB              & Average         \\ \hline
\textit{ST-EC-NER} (CE) & 0.0111          & 0.0167          & 0.0131          & 0.0146          & 0.0090          & 0.0132          & 0.0130          \\ \hline
\textit{ST-EC-NER} (TS) & 0.0100          & 0.0172          & 0.0136          & 0.0108          & 0.0087          & 0.0142          & 0.0124          \\ \hline
\textit{ST-EC-NER} (LS) & \textbf{0.0002} & \textbf{0.0002} & \textbf{0.0001} & \textbf{0.0002} & \textbf{0.0009} & \textbf{0.0010} & \textbf{0.0004} \\ \hline
\textit{ST-DA-NER} (CE) & 0.0090          & 0.0121          & 0.0117          & 0.0122          & 0.0082          & 0.0110          & 0.0107          \\ \hline
\multicolumn{8}{|c|}{}                                                                                                                         \\ \hline
\multicolumn{8}{|c|}{\textbf{DP + NER}}                                                                                                        \\ \hline
                 & BC              & BN              & MZ              & NW              & TC              & WB              & Average         \\ \hline
\textit{MT-AVG} (CE)    & 0.0436          & 0.0499          & 0.0473          & 0.0345          & 0.0393          & 0.1681          & 0.0637          \\ \hline
\textit{MT-AVG} (TS)    & 0.0447          & 0.0468          & 0.0460          & 0.0318          & 0.0396          & 0.1645          & 0.0622          \\ \hline
\textit{MT-AVG} (LS)    & \textbf{0.0040} & \textbf{0.0018} & \textbf{0.0013} & \textbf{0.0012} & \textbf{0.0057} & \textbf{0.0546} & \textbf{0.0114} \\ \hline
\textit{MT-AVGDA} (CE)    & 0.0291          & 0.0287          & 0.0302          & 0.0236          & 0.0257          & 0.0718          & 0.0349          \\ \hline
\end{tabular}
\end{adjustbox}
\caption{Overconfidence Error Results.}
\label{table:overconfidence}
\end{table}

\com{
\begin{figure}[!ht]
    \centering
    \subfloat[\centering Dependency Parsing]
    {{\includegraphics[width=0.5\textwidth,height=5cm]{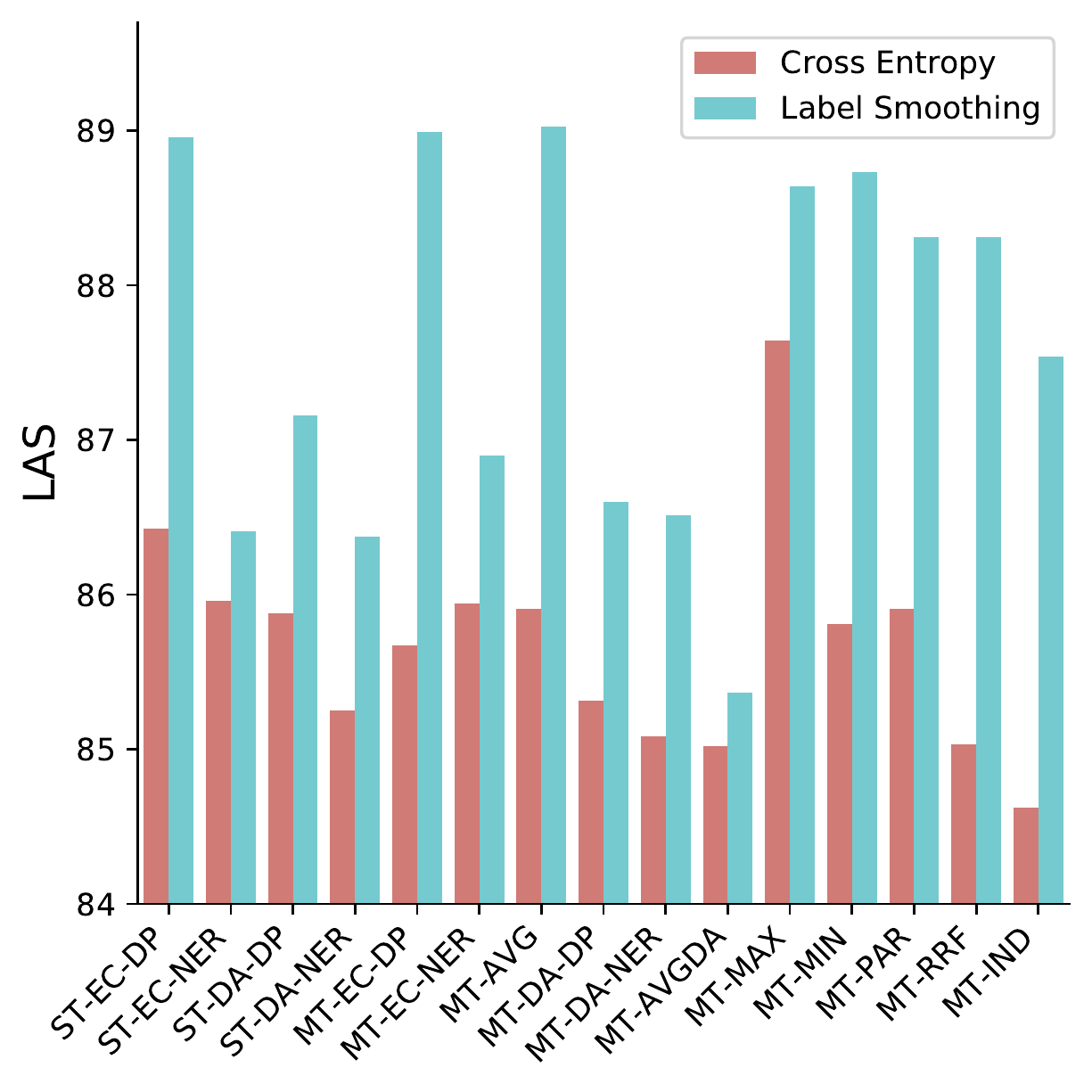}}}
    \qquad
    \centering
    \subfloat[\centering Named Entity Recognition]
    {{\includegraphics[width=0.5\textwidth,height=5cm]{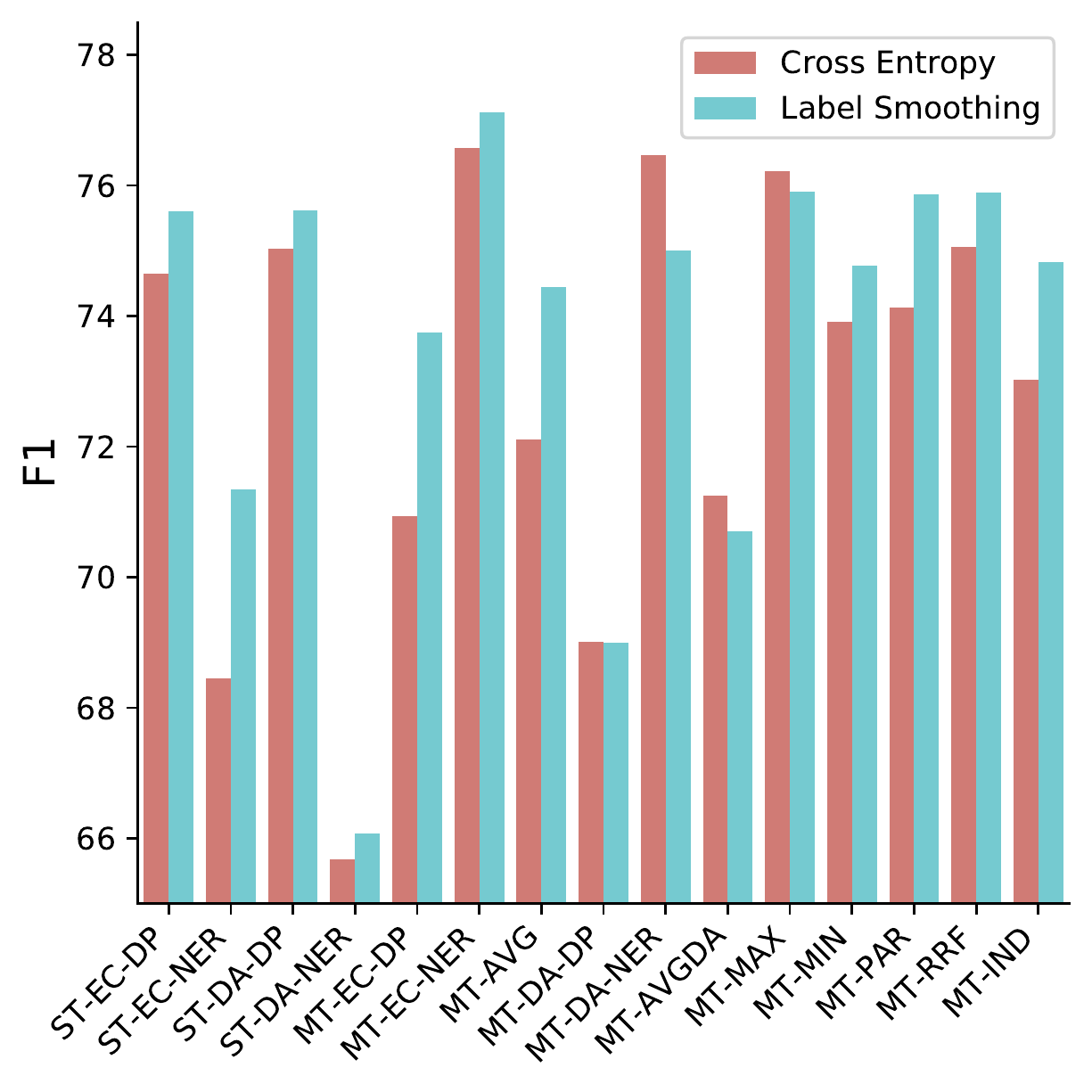}}}
    \caption{In-domain results: Average performance with cross-entropy vs. label smoothing for all AL methods (Q1.5).} 
    \label{fig:all_domains}
\end{figure}
}

\com{
\begin{figure}[!ht]
    \centering
    \subfloat[\centering Dependency Parsing]
    {{\includegraphics[width=0.5\textwidth,height=4cm]{figures/dp_st_vs_mt_best_cross_domain.png}}}
    \qquad
    \centering
    \subfloat[\centering Named Entity Recognition]
    {{\includegraphics[width=0.5\textwidth,height=4cm]{figures/ner_st_vs_mt_best_cross_domain.png}}}
    \caption{Cross-domain results: Best ST-AL vs. best MT-AL methods per domain (Q1.6).} 
    \label{fig:cross_domain_st_vs_mt}
\end{figure}
}

In Table \ref{table:overconfidence} we compare the $OE$ scores of \textit{ST-EC}, trained with the LS objective to 3 alternatives: \textit{ST-EC} trained with the CE objective, \textit{ST-EC} with the post-processing method \textit{temperature scaling} (TS), and \textit{ST-DA} trained with the CE objective. Both TS and dropout inference have been shown to improve confidence estimates \cite{guo2017calibration,ovadia2019can}, and hence serve as alternatives to LS in this comparison. $OE$ scores are reported on the unlabeled set (given the true labels in hindsight) at the final AL iteration for both tasks. Additionally, $OE$ scores for \textit{MT-AVG} and \textit{MT-AVGDA} are also reported and averaged on both tasks.

The results are conclusive, \textbf{\textit{LS is the least overconfident method, achieving the lowest OE scores on all 18 setups, but one}}. While LS achieves a proportionate reduction error (PRE) of between 57.2\% and 96.7\% compared to the standard CE method, DA achieves at most a PRE of 51.9\% and TS seems to have almost no effect. These results confirm that LS is highly effective in reducing overconfidence scores for BERT-based models, and we are able to show for the first time that such a reduction also holds for multi-task models. 

\com{
\paragraph{Label Smoothing vs. Cross-entropy Evaluation (Q1.5)}
After confirming that LS indeed reduces the overconfidence of our models, we now turn to ask whether it also improves the quality of the AL methods.
Figure \ref{fig:all_domains} presents the AL results at the final AL iteration for all confidence scores when trained with the LS or CE objectives. 

The results suggest that LS is very effective for DP, outperforming CE in all comparisons, with an average improvement of 1.9\%. For NER, LS is mainly effective for the entropy-based methods, except for \textit{MT-MAX}, while CE is found effective for MT dropout-based methods. The lower impact of LS on NER may suggest that when CE $OE$ is small to begin with (Table \ref{table:overconfidence}), further reduction with LS does not affect AL, even when the OE reduction is substantial.}

\com{Finally, we report that in more than 77\% of the experiments LS is better than CE (57\% of all experiments are statistically significant).}

\com{
For the DP task LS is less effective for single-task random selection and for NER-based selection, whereas, for NER, LS is less effective for single-task random selection and for joint-task selection (\textit{MT-AVG}, \textit{MT-AVGDA}, \textit{MT-MAX}). However, when considering all methods, LS is more effective than CE for exactly twice of the different AL methods, 8 out of the 12, in both DP and NER tasks. Improvements reach up to 0.74\% LAS and 1.81\% F1 for DP and NER, respectively.

Furthermore, we perform CE and LS training with all AL selection methods on each of the OntoNotes domains separately. The domains that most benefit from LS are BC for DP with an average improvement of 2.56\% LAS and TC for NER with an average improvement of 2.23\% F1.
When averaging all experiments, LS is better than CE in 0.52\% LAS and 1.66\% F1 for DP and NER, respectively.

These results confirm the first part of Q1.5, LS is beneficial for AL both for single- and multi-task learning and on average is better than CE.}

\com{
\begin{figure}[!ht]
    \centering
    \subfloat[\centering Dependency Parsing]
    {{\includegraphics[width=0.5\textwidth]{figures/cross_domain_heatmap_dp_crop.png}}}
    \qquad
    \centering
    \subfloat[\centering Named Entity Recognition]
    {{\includegraphics[width=0.5\textwidth]{figures/cross_domain_heatmap_ner_crop.png}}}
    \caption{Cross-domain results: Performance difference between LS and CE (Q1.6).} 
    \label{fig:cross_domain_heatmap}
\end{figure}

\paragraph{Cross-domain Results (Q1.6-Q1.7)}

Finally, we aim to evaluate the different models in cross-domain performance. We ground this analysis in two important comparisons: multi-task vs. single-task performance and LS vs. CE performance. Following the setup of \citet{rotman2019deep}, we set the news domain (NW), the largest OntoNotes domain, as our source domain and treat all other domains as targets. 

Before delving into our comparisons, we ask whether AL is effective in our cross-domain, zero-shot setup. While this is not demonstrated in graphs or tables, random selection is indeed outperformed by the more sophisticated AL methods: For DP, \textit{ST-R} is on average 0.89\% LAS worse than the best ST-AL method, \textit{ST-EC-DP}, and \textit{MT-R} is on average 1.63\% LAS worse than the best MT-AL method, \textit{MT-MAX}. In NER, \textit{ST-R} and \textit{MT-R} are on average 5.2 and 2 F1 points worse than the best performing methods, \textit{ST-EC-ER} and \textit{MT-EC-NER}, respectively.

Figure \ref{fig:cross_domain_st_vs_mt} compares for each target domain its best ST-AL method with its best MT-AL method, both for DP and NER, in the final AL iteration. In all cases, the best models employ the LS objective (see Q1.7 for a detailed comparison between LS and CE for cross-domain generalization). Overall, the best MT-AL methods outperform the best ST-AL methods for all target domains in both tasks, with the only exception of TC in NER (5 of the 10 comparisons are statistically significant). 

We finalize our answer to Q1.6 by noticing that the average performance gaps between ST-AL and MT-AL are not big (0.56\% LAS and 0.33 F1 in favor of MT-AL), and hence concluding that MT-AL methods are somewhat effective for zero-shot performance on unseen domains.


In order to answer Q1.7 Figure \ref{fig:cross_domain_heatmap} plots the performance gaps between LS and CE on both tasks for all target domains, considering the last iteration of the AL method. The figure indicates that LS is effective for cross-domain DP performance, outperforming the CE objective in all experiments but 4, with an average improvement of 0.66\% LAS. LS is also beneficial for NER, achieving an average improvement of 1.17 F1 points, where in six cases the improvement ranges between 5.35 and 11.04 F1 points. The difference is statistically significant in 46.6\% of the comparisons.

\com{The figure indicates that LS is highly beneficial for cross-domain DP performance by surpassing the CE objective in all experiments but 4, corresponding to either the TC domain or the \textit{MT-MAX} method. On average, LS is better than CE for all domains and for all AL methods, but \textit{MT-MAX}, with a total average improvement of 0.66\% LAS.

Similar patterns also occur for NER, where in two cases LS improves CE in approximately 5.5\% F1, and in four cases it improves between 9.19\% and 11.04\% F1. Again, On average LS is better than CE for all domains and for all AL methods, but two, with a total average improvement of 1.17\% F1.}

Overall, this comparison concludes our final question about the impact of LS on AL. We extend the observation of \citet{desai2020calibration} who showed that LS is effective in reducing overconfidence in out-of-domain predictions and confirm that LS is also capable of delivering better AL performance in this setup. Following our answers to Q1.4 and Q1.5, we now conclude that LS is effective in both in- and cross-domain setups: Both for overconfidence reduction and for improved AL.
}

\section{MT-AL for Hierarchically-related Tasks}
\label{sec:hierarchical_tasks}

Until now, we have considered tasks (DP and NER) that are mutually informative but can be trained independently of each other. However, other multi-task learning scenarios involve a task that is dependent on the output of another task. A prominent example is the relation extraction (RE) task that depends on the output of the NER task, since the goal of RE is to classify and identify relations between named entities. Importantly, if the NER part of the model does not perform well, this harms the RE performance as well. Sample selection in such a setup should hence reflect the hierarchical relation between the tasks.

\subsection{Selection Methods}

Since the quality of the classifier for the independent task (NER) now affects also the quality of the classifier for the dependent task (RE), the confidence of each of the tasks may get different relative importance values. Although this in principle can also be true for independent tasks (\S \ref{sec:closely_related_tasks}), explicitly accounting for this property seems more crucial in the current setup.

We hence modify four of our joint-selection methods ($\S$ \ref{sec:al_methods}) to reflect the inherent a-symmetry between the tasks, by presenting a scaling parameter $0\leq \beta \leq 1$: \footnote{\label{fn:equal_weights} \guyfix{We do not include the \textit{MT-MAX} and \textit{MT-MIN} methods in our evaluation. Since the two tasks exhibit confidence scores in a similar range, scaling their confidence scores according to these methods resulted in selecting samples based almost solely on the task with the higher (in the case of \textit{MT-MAX}) or lower (in the case of \textit{MT-MIN}) scaling parameter.}}\\
%
\textbf{a)} \textit{MT-AVG} is now calculated as follows:
\begin{align*}
    \textit{MT-AVG}(x) &= \beta \cdot \textit{MT-EC-RE}(x) + \\
    & \quad (1-\beta) \cdot \textit{MT-EC-NER}(x).
\end{align*}
\textbf{b)} $\textit{MT-RRF}$ is calculated similarly by multiplying the RRF term of RE by $\beta$ and that of NER by $1-\beta$. \\
\textbf{c)} $\textit{MT-IND}$ is calculated by independently choosing $100\cdot\beta \%$ of the selected samples according to the RE scores and $100\cdot(1 - \beta) \%$ according to the NER scores.\\
\textbf{d)} \textit{MT-PAR} is computed by restricting the first Pareto condition for the position of the RE confidence score: $c_{RE} \leq q_{\beta} \cdot c'_{RE}$, where $q_{\beta}$ is the value of the $\beta$-quantile of the RE confidence scores. \footnote{\guyfix{The second Pareto condition is similarly modified, but now with the $<$ sign.}} \guyfix{We apply such a condition if $\beta < 0.5$. Otherwise, if $\beta > 0.5$ the condition is applied to the NER component, and when it is equal to $0.5$, the original Pareto method is used.} Since we restrict the condition to only one of the tasks, fewer samples will meet this condition (since $0 \leq q_{\beta} \leq 1$), and the Pareto frontier will include more samples that have met the condition for the second task.

\subsection{Research Questions}

In our experiments, we would like to explore two research questions:
\textbf{Q2.1}: Which MT-AL selection methods are most suitable for this setup ? and
\textbf{Q2.2}: What is the best balance between the participating tasks ?

Since RE fully relies on the output of NER, we limit our experiments only to joint multi-task models and do not include single-task models.

\subsection{Experimental Setup}

We experiment with the span-based joint NER and RE BERT model of \citet{li-etal-2021-weakly}.\footnote{\url{https://github.com/JiachengLi1995/JointIE}.} Experiments were conducted on five diverse datasets: NYT24 and NYT29 \cite{nayak2020effective}, ScieRC \cite{luan-etal-2018-multi}, WebNLG \cite{gardent-etal-2017-webnlg}, and WLP \cite{kulkarni-etal-2018-annotated}. The AL setup is identical to that of \S \ref{sec:exp-setup1}. Other hyper-parameters that were not mentioned before are identical to those of the original implementation.
\com{At each AL iteration, we set the training steps to 20K with the early stopping criteria,\footnote{Roi: I do not understand the text from the beginning of the sentence till here. \guyfix{we performed 20K training (gradient) step updates in every AL iteration and performed an early stopping according to the dev set. Anyway, I removed these lines since they are identical to what was written in Section 5.4.}}1,000 warm-up steps and LS of $\alpha=0.2$. Other hyper-parameters are identical to those of the original implementation.}

\subsection{Results}

\paragraph{Best Selection Method (Q2.1)}

We start by identifying the best selection method for this setup. Table \ref{table:hierarchical_methods_results} summarizes the per-task average score for the best $\beta$ value of each method across the five datasets.

We observe three interesting patterns. First, \textbf{\textit{MT-AL is very effective in this setup for the dependent task (RE)}}, while for the independent task (NER), random selection does not fall too far behind. 
Second, \textbf{\textit{all MT-AL methods achieve better performance for higher}} $\bm{\beta}$ \textbf{\textit{values}} by giving more weight to the RE confidence scores during the selection process. This is an indication that indeed the selection method should reflect the asymmetric nature of the tasks. \guyfix{Third, overall, \textbf{\textit{\textit{MT-IND} is the best performing method}}, averaging first in NER and in RE, while \textit{MT-AVG}, \textit{MT-RRF} and \textit{MT-PAR} achieve similar results in both tasks.} 

\com{
\begin{table}[t!]
\centering
\begin{tabular}{|c|c|c|}
\hline
       & NER            & RE             \\ \hline
MT-R   & 81.96          & 52.47          \\ \hline
MT-AVG & 82.69          & \textbf{61.88} \\ \hline
MT-RRF & 82.56          & 58.68          \\ \hline
MT-IND & 82.94          & 60.29          \\ \hline
MT-PAR & \textbf{82.95} & 60.66          \\ \hline
\end{tabular}
\caption{Hierarchical MT-AL results. We report best average F1 results over all five datasets.}
\label{table:hierarchical_methods_results}
\end{table}}

\begin{table}[t!]
\centering
\begin{adjustbox}{width=\linewidth}
\begin{tabular}{|c|c|c|c|c|}
\hline
                     & NER                             & NER Best & RE                              & RE Best \\ \hline
\textit{MT-R}                 & 81.96                           & 0/5      & 52.47                           & 05      \\ \hline
\textit{MT-AVG} ($\beta=1.0$)   & 82.64                           & 1/5      & 59.51                           & 2/5     \\ \hline
\textit{MT-RRF} ($\beta=0.8$) & 82.69                           & 1/5      & 59.65                           & 1/5     \\ \hline
\textit{MT-IND} ($\beta=1.0$)   & \textbf{82.86} & 2/5      & \textbf{60.15}                           & 0/5     \\ \hline
\textit{MT-PAR} ($\beta=0.7$) & 82.68                           & 1/5      & 59.57 & 2/5     \\ \hline
\end{tabular}
\end{adjustbox}
\caption{Hierarchical MT-AL results. We report best average F1 results over all five datasets for the best $\beta$ configuration per method.}
\label{table:hierarchical_methods_results}
\end{table} 

\paragraph{Scaling Configuration (Q2.2)}

\begin{figure}[!t]
    \centering
    \subfloat[\centering ScieRC]
    {{\includegraphics[width=\linewidth]{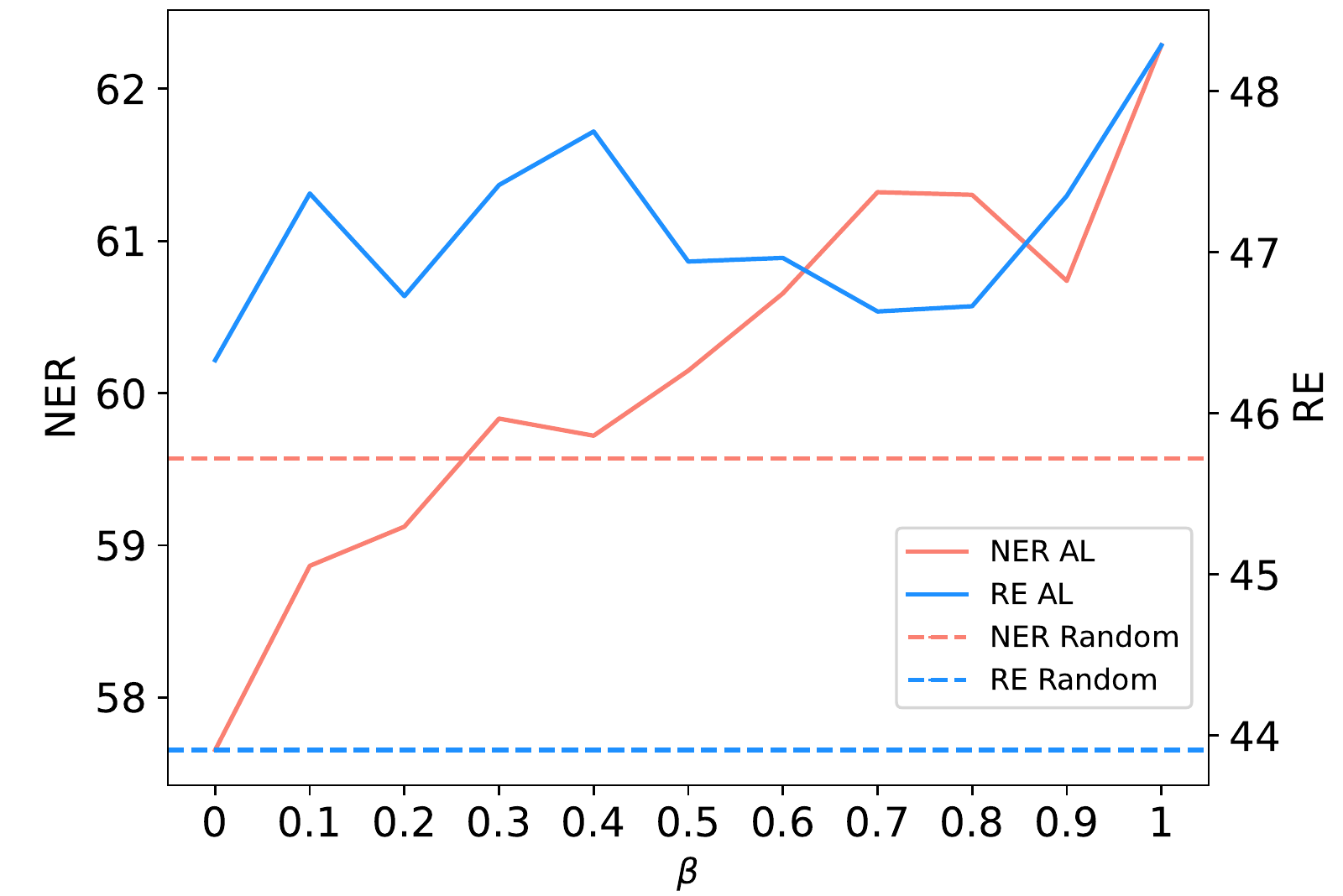}}}
    \qquad
    \centering
    \subfloat[\centering WebNLG]
    {{\includegraphics[width=\linewidth]{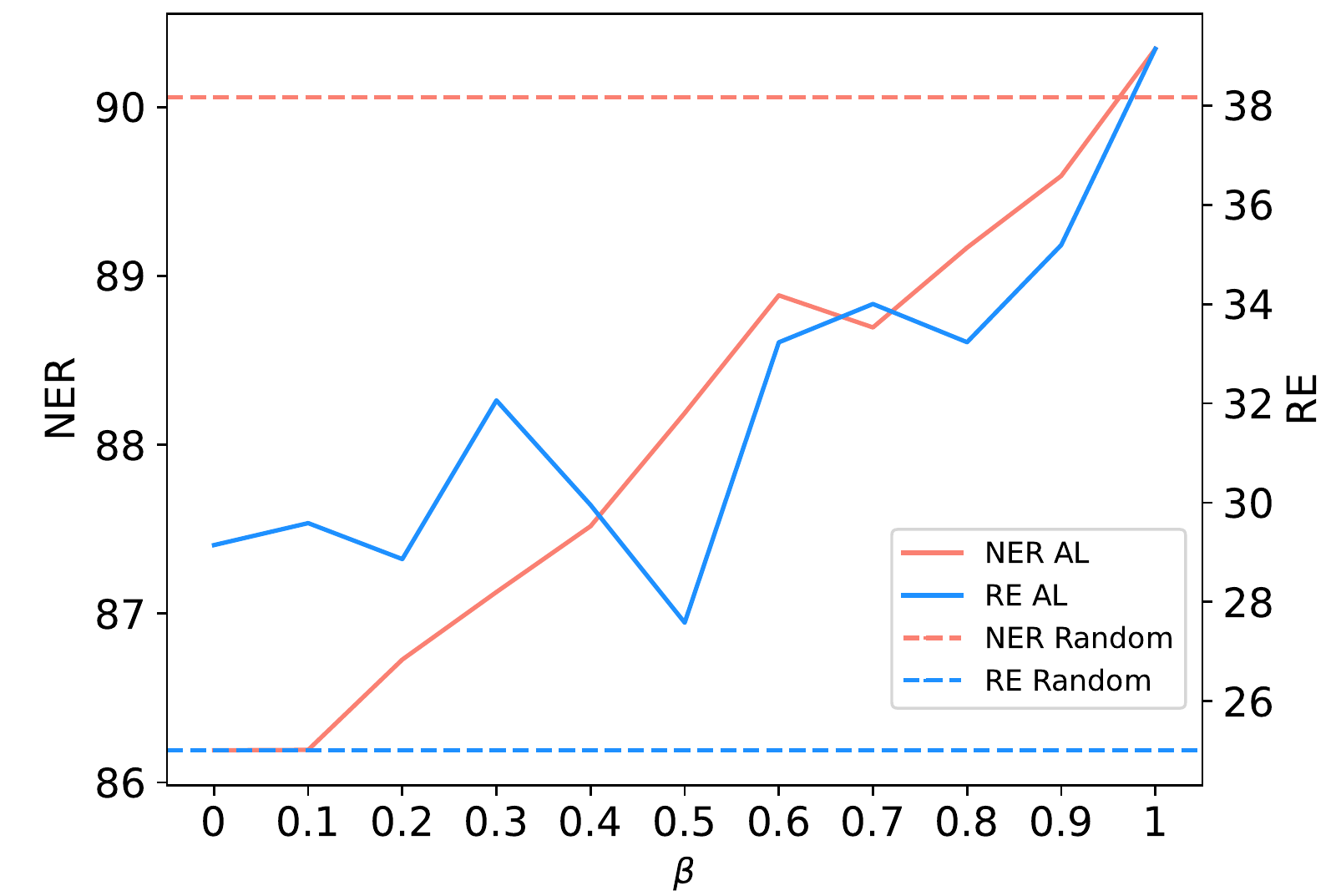}}}
    \caption{Average F1 scores over four joint-selection methods as a function of $\beta$ (the relative weight of the RE confidence).} 
    \label{fig:ner_re_avg_beta}
\end{figure}

Figure \ref{fig:ner_re_avg_beta} presents the average F1 scores of the four joint-selection methods, as well as the random selection method \textit{MT-R}, as a function of $\beta$ (the relative weight of the RE confidence). First, we notice that joint selection outperforms random selection in the WebNLG domain only for RE (the dependent task) but not for NER (except when $\beta$ approaches $1$). Second, and more importantly, $\beta=1$, that is, \textbf{\textit{selecting examples only according to the confidence score of the RE (dependent) task, is most beneficial for both tasks}} (Q2.1). We hypothesize that this stems from the fact that the RE confidence score conveys information about both tasks and that this combined information provides a stronger signal with respect to the NER (independent) task, compared to the NER confidence score. Interestingly, the positive impact of higher values of $\beta$ is more prominent for NER, even though this means that the role of the NER confidence score is downplayed in the sample selection process.

\guyfix{For the individual selection methods, we report that \textit{MT-AVG} and \textit{MT-IND} achieve higher results as $\beta$ increases, while \textit{MT-PAR} and \textit{MT-RRF} peak at $\beta=0.7$ and $\beta=0.8$, respectively, and then drop by 0.2 F1 points for NER and 0.9 F1 points for RE on average.}
%
%
%

\com{
\begin{figure*}[!h]
\centering
\includegraphics[width=0.99\linewidth]{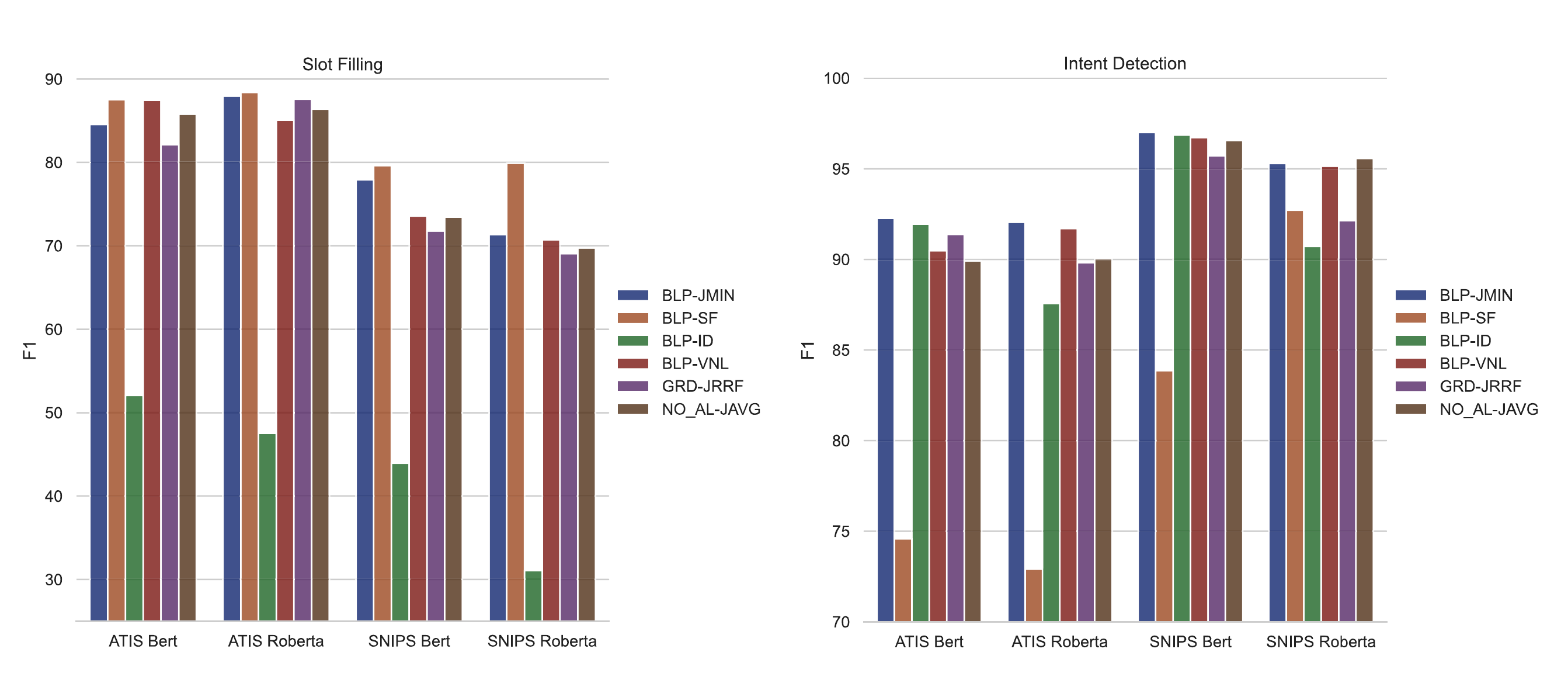}
\caption{Cost-sensitive MT-AL performance (Q3.2).}
\label{fig:slot_intent}
\end{figure*}}

\section{MT-AL for Tasks with Different Annotation Granularity}
\label{sec:cost_sensitive_results}

NLP tasks are defined on different textual units, with the most common examples of sentence-level and token-level tasks. Our last investigation considers the scenario of two closely-related tasks that are of different granularity: Slot filling (SF, token-level) and intent detection (ID, sentence-level).
%

Due to the different annotation nature of the two tasks, we have to define the cost of example annotation with respect to each. Naturally, there is no correct way to quantify these costs, but we aim to propose a realistic model. We denote the cost of annotating a sample for SF with $Cost_{SF} = m + tp\cdot nt$, where $m$ is the number of tokens in the sentence, $tp$ is a fixed token annotation cost (we set $tp=1$) and $nt$ is the number of entities. The cost of annotating a sample for ID is next denoted with $Cost_{ID} = m + ts$, where $ts$ is a fixed sentence cost (we set $ts=3$). Our solution (see below) allows some examples to be annotated only with respect to one of the tasks. For examples that are annotated with respect to both tasks we consider an additive joint cost where the token-level term $m$ is considered only once: $JCost = Cost_{SF} + Cost_{ID} - m$. In our experiments, we allow a fixed annotation budget of $B=500$ per AL iteration.


\subsection{Methods}

We consider three types of decision methods: \textbf{Greedy Active Learning (GRD\_AL)}: AL methods that at each iteration greedily choose the least confident samples until the budget limitation is reached; \textbf{Binary Linear Programming Active Learning (BLP\_AL)}: AL methods that at each iteration opt to minimize the sum of confidence scores of the chosen samples given the budget constraints. The optimization problem (see below) is solved using a BLP solver;\footnote{\url{https://www.python-mip.com/}.} and \textbf{Binary Linear Programming (BLP)}: an algorithm that after training on the initial training set chooses all the samples at once, by solving the same constrained optimization problem as in BLP\_AL.


For each of these categories, we experiment with four families of AL methods: \textbf{a) Unrestricted Disjoint Sets (UDJS):} This selection method is based on the non-aggregated multi-task confidence scores, where each sample can be chosen to be annotated on either task or both. The UDJS optimization problem aims to maximize the uncertainty scores ($1 - Conf_t(x)$) of the selected samples given the budget and selection constraints:

\begin{maxi*}|s|
{}{\sum_{x \in \textit{U}} \sum_{t \in \textit{T}} (1 - Conf_t(x)) \cdot X_t(x)}
{}{}
\addConstraint{\sum_{x \in \textit{U}} \sum_{t \in \textit{T}} Cost_t(x) \cdot (X_t(x)-Y(x))
\\&& + JCost(x) \cdot Y(x) \leq B}
\addConstraint{\frac{1}{|\textit{T}|}\sum_{t \in \textit{T}} X_t(x) \geq Y(x) \quad \forall x \in \textit{U}}
\addConstraint{X_t(x), Y(x) \in \{0,1\} \quad \forall x \in \textit{U}, t \in \textit{T}},
{}
\end{maxi*}

where \textit{U} is the unlabeled set, \textit{T} is the set of tasks, $Conf_t$ is the \textit{MT-EC-t} confidence score, $X_t(x)$ is a binary indicator indicating the annotation of sample $x$ on task $t$, and $Y(x)$ is a binary indicator indicating the annotation of $x$ on all tasks. 

Notice that this formulation may yield annotated examples for only one of the tasks, although this is unlikely, particularly under an iterative protocol when the confidence scores of the models are updated after each iteration. 

\textbf{b) Equal Budget Disjoint Sets (EQB-DJS)}: This strategy is similar to the above UDJS except that the budget is equally divided between the two tasks and the optimization problem is solved for each of them separately. If a sample is chosen to be annotated for both tasks, we update its cost according to the joint cost and re-solve the optimization problems until the entire budget is used. 

\textbf{c-f) Joint-task Selection}: A sample could only be chosen to be annotated on both tasks, where confidence scores are calculated using a multi-task aggregation. The BLP optimization problem is formulated as follows:

\begin{maxi*}|s|
{}{\sum_{x \in \textit{U}} (1 - Conf(x)) \cdot Y(x)}
{}{}
\addConstraint{\sum_{x \in \textit{U}} JCost(x) \cdot Y(x) \leq B}
\addConstraint{Y(x) \in \{0,1\} \quad \forall x \in \textit{U}},
{}
\end{maxi*}
where $Conf$ is calculated by \textbf{c) \textit{MT-AVG}}, \textbf{d) \textit{MT-MAX}}, \textbf{e) \textit{MT-MIN}} or \textbf{f) \textit{MT-RRF}}. \footnote{\label{fn:greedy_selection}\guyfix{Since \textit{MT-PAR} and \textit{MT-IND} do not score the examples, they can only be applied with the greedy decision method. A discussion on their results is provided in $\S$ \ref{sec:overall_comparison}.}}

\textbf{g-j) Single-task Confidence Selection (STCS)}: A sample could only be chosen to be annotated on both tasks, where the selection process aims to maximize the uncertainty scores of only one of the tasks: \textbf{g) STCS-SF} or \textbf{j) STCS-ID}. Similarly to Joint-task Selection, the budget constraints are applied to the joint costs.

\com{\textbf{g-j) Single-task Selection}: samples are only annotated on one of the tasks according to the \textit{ST-EC} scores. We experiment both using a multi-task model: \textbf{g) MT-SF} - annotating only SF labels, \textbf{h) MT-ID} - annotating only ID labels, or using a single-task model \textbf{i) ST-SF} and \textbf{j) ST-ID}. As a convention, we will denote the decision method with a standard script and its AL method in subscript.\footnote{Roi: I do not understand what this family of methods is. If you sample only with respect to one task, how do you have annotations for the other task? \guyfix{I only have the initial labeled set that includes annotations for both task and then I add only annotations for one of the tasks.}}}

\subsection{Research Questions}

We focus on three research questions: \textbf{Q3.1}: Does BLP optimization improve upon greedy selection?
\textbf{Q3.2}: Do optimization selection and active learning have a complementary effect? and \textbf{Q3.3}: Is it better to annotate all samples on both tasks or to construct a disjoint annotated training set? 

\subsection{Experimental Setup}

We conduct experiments on two prominent datasets: ATIS \cite{price1990evaluation} and SNIPS \cite{coucke2018snips}, and consider two Transformer-based encoders: BERT-base \cite{devlin2019bert} and Roberta-base \cite{liu2019roberta}. Our code is largely based on the implementation of \citet{zhu2017encoder}.\footnote{\url{https://github.com/sz128/slot_filling_and_intent_detection_of_SLU}.} We run the AL process for 5 iterations with an initial training set of 50 random samples and a fixed-size development set of 100 random samples. We train all models for 30 epochs per iteration, with an early stopping criterion and with label smoothing ($\alpha=0.1$). Other hyper-parameters were set to their default values. 

\begin{table}[!t]
\centering
\begin{adjustbox}{width=\linewidth}
\begin{tabular}{|c|cccc|cccc|}
\hline
            & \multicolumn{4}{c|}{ATIS}                                                                                                        & \multicolumn{4}{c|}{SNIPS}                                                                                                   \\ \hline
            & \multicolumn{2}{c|}{BERT}                                                 & \multicolumn{2}{c|}{Roberta}                         & \multicolumn{2}{c|}{BERT}                                             & \multicolumn{2}{c|}{Roberta}                         \\ \hline
            & \multicolumn{1}{c|}{SF}             & \multicolumn{1}{c|}{ID}             & \multicolumn{1}{c|}{SF}             & ID             & \multicolumn{1}{c|}{SF}            & \multicolumn{1}{c|}{ID}          & \multicolumn{1}{c|}{SF}             & ID             \\ \hline
\textit{MT-MIN}$_{BLP\_AL}$ & \multicolumn{1}{c|}{84.53}          & \multicolumn{1}{c|}{\textbf{92.27}} & \multicolumn{1}{c|}{\textbf{87.94}} & \textbf{92.05} & \multicolumn{1}{c|}{\textbf{77.90}} & \multicolumn{1}{c|}{\textbf{97.00}} & \multicolumn{1}{c|}{\textbf{71.35}} & 95.29          \\ \hline
\textit{MT-RRF}$_{GRD\_AL}$ & \multicolumn{1}{c|}{82.11}          & \multicolumn{1}{c|}{91.38}          & \multicolumn{1}{c|}{87.57}          & 89.81          & \multicolumn{1}{c|}{71.78}         & \multicolumn{1}{c|}{95.71}       & \multicolumn{1}{c|}{69.06}          & 92.14          \\ \hline
\textit{MT-AVG}$_{BLP}$     & \multicolumn{1}{c|}{\textbf{85.78}} & \multicolumn{1}{c|}{89.92}          & \multicolumn{1}{c|}{86.38}          & 90.03          & \multicolumn{1}{c|}{73.46}         & \multicolumn{1}{c|}{96.57}       & \multicolumn{1}{c|}{69.73}          & \textbf{95.57} \\ \hline
\end{tabular}
\end{adjustbox}
\caption{Comparison of decision categories (Q3.1 and Q3.2).}
\label{table:slot_intent_main_results}
\end{table}

\begin{table}[!t]
\centering
\begin{adjustbox}{width=\linewidth}
\begin{tabular}{|c|cccc|cccc|}
\hline
            & \multicolumn{4}{c|}{ATIS}                                                                                                                                                        & \multicolumn{4}{c|}{SNIPS}                                                                                                   \\ \hline
            & \multicolumn{2}{c|}{BERT}                                                                         & \multicolumn{2}{c|}{Roberta}                                                 & \multicolumn{2}{c|}{BERT}                                             & \multicolumn{2}{c|}{Roberta}                         \\ \hline
            & \multicolumn{1}{c|}{SF}             & \multicolumn{1}{c|}{ID}                                     & \multicolumn{1}{c|}{SF}             & ID                                     & \multicolumn{1}{c|}{SF}            & \multicolumn{1}{c|}{ID}          & \multicolumn{1}{c|}{SF}             & ID             \\ \hline
\textit{MT-MIN}$_{BLP\_AL}$ & \multicolumn{1}{c|}{84.53}          & \multicolumn{1}{c|}{\textbf{92.27}} & \multicolumn{1}{c|}{\textbf{87.94}} & \textbf{92.05} & \multicolumn{1}{c|}{\textbf{77.90}} & \multicolumn{1}{c|}{\textbf{97.00}} & \multicolumn{1}{c|}{\textbf{71.35}} & \textbf{95.29} \\ \hline
\textit{UDJS}$_{BLP\_AL}$ & \multicolumn{1}{c|}{\textbf{87.46}} & \multicolumn{1}{c|}{90.48}          & \multicolumn{1}{c|}{85.05}          & 91.71          & \multicolumn{1}{c|}{73.56}         & \multicolumn{1}{c|}{96.71}       & \multicolumn{1}{c|}{70.72}          & 95.14          \\ \hline
\textit{EQB-DJS}$_{BLP\_AL}$ & \multicolumn{1}{c|}{86.59} & \multicolumn{1}{c|}{88.02}          & \multicolumn{1}{c|}{85.25}          & 89.14          & \multicolumn{1}{c|}{69.92}         & \multicolumn{1}{c|}{91.43}       & \multicolumn{1}{c|}{63.58}          & 95.00             \\ \hline
\end{tabular}
\end{adjustbox}
\caption{Comparison of joint-task selection methods (Q3.3).}
\label{table:slot_intent_q3}
\end{table}

\subsection{Results}
\paragraph{Optimization-based Selection and AL (Q3.1 and Q3.2)}

To answer the first two questions, we show in Table \ref{table:slot_intent_main_results} the final sample F1 performance for both tasks, when selection is done with the best selection method of each decision category: \textit{MT-RRF}$_{GRD\_AL}$, \textit{MT-MIN}$_{BLP\_AL}$ and \textit{MT-AVG}$_{BLP}$. The results confirm that \textbf{\textit{the BLP optimization (with AL) is indeed superior to greedy AL selection}}, surpassing it in all setups (two tasks, two datasets, and two pre-trained language encoders, 5 of the comparisons are statistically significant).

\textbf{\textit{The answer to Q3.2 is also mostly positive.}} BLP optimization and iterative AL have a complementary effect, as \textit{MT-MIN}$_{BLP\_AL}$ in most cases achieves higher performance than \textit{MT-AVG}$_{BLP}$ (50\% of the results are statistically significant). 

\guyfix{In fact, \textbf{\textit{the answer for the two questions holds true for all selection methods}}, as all perform best using our novel \textbf{BLP\_AL} decision method. Second is \textbf{BLP}, indicating that the BLP formulation is highly effective for MT setups under different budget constraints. Finally, last is the standard greedy procedure (\textbf{GRD\_AL}) which is commonly used in the AL literature.} 

%

\paragraph{Joint-task Selection (Q3.3)}

To answer Q3.3, we perform a similar comparison in Table \ref{table:slot_intent_q3}, but now for the most prominent methods for each joint-task selection method. 
The results indicate that \textit{MT-MIN}$_{BLP\_AL}$ that enforces \textbf{\textit{joint annotation is better than allowing for non-restricted disjoint annotation}} (\textit{UDJS}$_{BLP\_AL}$) \textbf{\textit{and than equally splitting the budget between the two tasks}} (\textit{EQB-DJS}$_{BLP\_AL}$), where 9 of the 16 comparisons are statistically significant.
We hypothesize that the superiority of \textit{MT-MIN}$_{BLP\_AL}$ stems from two main reasons: 1. Integrating the joint aggregation confidence score into the optimization function provides a better signal than the single-task confidence scores; 2. Having a joint dataset where all samples are annotated on both tasks rather than a disjointly annotated (and larger) dataset allows the model to achieve better performance since the two tasks are closely related.

Finally, we also report that ST-AL experiments have led to poor performance. The ST-AL methods trail by 8.8 (SF) and 4.7 (ID) F1 points from the best MT-AL method \textit{MT-MIN}$_{BLP\_AL}$ on average. Interestingly, we also observe that selection according to ID (\textit{STCS-ID}$_{BLP\_AL}$) has led to better average results on both tasks than selection according to SF (\textit{STCS-SF}$_{BLP\_AL}$), suggesting that similarly to $\S$ \ref{sec:hierarchical_tasks}, selection according to the higher-level task often yields better results.

\com{another indication of the need to construct a joint annotated data set that comes from the direct comparison of MT-AL vs. ST-AL, as studied similarly in $\S$ \ref{sec:closely_related_tasks}. In this setup, we observe similar patterns and report that although the methods that base their selection only on one of the tasks (\textit{ST-SF}, \textit{ST-ID}, \textit{MT-SF}, and \textit{MT-ID}) were prominent for in-task performance, these methods have performed very poorly when evaluating the model on the other task, the one that was not included in the selection process.}

\com{
The joint-task selection methods using the BLP solver are the best models overall, with BLP-MIN reaching the highest average score of 86.02 (averaging on both tasks, the two datasets, and the two encoder models). Such a result derives three main conclusions: a BLP solver running offline at the end of each AL iteration is better than greedy or non-iterative optimizations for MT-AL under budget constraints (Q3.1). Second, incorporating an aggregation of the single-task uncertainties (using \textit{MT-MIN} in most cases) is better than a vanilla, an equal budget, or a single-task selection approach (Q3.2). Thirdly, under our cost assumptions, annotating all samples on both tasks is more valuable than other annotation approaches (Q3.3). Figure \ref{fig:slot_intent} additionally confirms this final conclusion, as it can be seen that the single-task selection methods (BLP-SF and BLP-ID) are prominent for in-task performance, however, they suffer from poor cross-task performance. Finally, we also note that we observed similar patterns in the \textbf{Multiplicative} joint-cost setup; however, most baselines have gained better results in the Additive setup, which by definition favors the selection of joint-task annotation, while \textbf{Multiplicative} favors the selection of single-task annotation for high values of $\gamma$ (usually greater than 0.85).}

\section{Overall Comparison}
\label{sec:overall_comparison}

\begin{table*}[!ht]
\centering
\begin{tabular}{|c|cc|cc|cc|c|}
\hline
       & \multicolumn{2}{c|}{DP + NER}                        & \multicolumn{2}{c|}{NER + RE}                        & \multicolumn{2}{c|}{SF + ID}                   &                \\ \hline
       & \multicolumn{1}{c|}{DP}             & NER            & \multicolumn{1}{c|}{NER}            & RE             & \multicolumn{1}{c|}{SF}           & ID         & Average        \\ \hline
\textit{MT-R}   & \multicolumn{1}{c|}{86.40}          & 70.78          & \multicolumn{1}{c|}{81.96}          & 52.47          & \multicolumn{1}{c|}{76.94}          & 91.95          & 76.75          \\ \hline
\textit{MT-AVG} & \multicolumn{1}{c|}{\textbf{89.03}} & 74.44          & \multicolumn{1}{c|}{80.23}          & 55.97          & \multicolumn{1}{c|}{76.42}          & 91.09          & 77.86          \\ \hline
\textit{MT-MAX} & \multicolumn{1}{c|}{88.64}          & \textbf{75.90} & \multicolumn{1}{c|}{81.50}          & \textbf{57.74} & \multicolumn{1}{c|}{76.64}          & 89.17          & 78.27          \\ \hline
\textit{MT-MIN} & \multicolumn{1}{c|}{88.73}          & 74.77          & \multicolumn{1}{c|}{81.32}          & 54.76          & \multicolumn{1}{c|}{73.75}          & 91.74          & 77.51          \\ \hline
\textit{MT-PAR} & \multicolumn{1}{c|}{88.31}          & 75.86          & \multicolumn{1}{c|}{81.47}          & 56.92          & \multicolumn{1}{c|}{\textbf{77.72}} & 90.28          & \textbf{78.43} \\ \hline
\textit{MT-RRF} & \multicolumn{1}{c|}{88.31}          & 75.89          & \multicolumn{1}{c|}{81.78}          & 54.29          & \multicolumn{1}{c|}{77.63}          & \textbf{92.26} & 78.36          \\ \hline
\textit{MT-IND} & \multicolumn{1}{c|}{87.54}          & 74.83          & \multicolumn{1}{c|}{\textbf{82.13}} & 54.85          & \multicolumn{1}{c|}{76.67}          & 91.44          & 77.91          \\ \hline
\end{tabular}
\caption{Average results of the joint multi-task selection methods across all three setups.}
\label{table:ablation_mt_selection_methods}
\end{table*}

\guyfix{As a final evaluation, we turn to compare the performance of our proposed joint-selection MT-AL methods in all setups. In our first setup (\S \ref{sec:closely_related_tasks}) we implemented all MT-AL selection methods with greedy selection, considering uniform task importance. Thereafter (\S \ref{sec:hierarchical_tasks} and \S \ref{sec:cost_sensitive_results}), we showed how these selection methods can be modified in the next two setups by either integrating non-uniform task weights or replacing the greedy selection with a BLP loss function overconfidence scores. However, not all of our MT-AL selection methods can be modified in such ways. \footnote{See Footnotes \ref{fn:equal_weights} and \ref{fn:greedy_selection} for further details.} We hence report our final comparison given the conditions applied in the first setup: Assuming uniform task weights and selecting samples in a greedy manner. Table \ref{table:ablation_mt_selection_methods} summarizes the average performance of the MT-AL methods in our three proposed setups: Complementing tasks (DP + NER), hierarchically-related tasks (NER + RE) and tasks of different annotation granularity (SF + ID).
}

\guyfix{Each selection method has its cons and pros, which we outline in our final discussion:\\
\textbf{\textit{MT-R}}, not surprisingly, is the worst performing method on average, as it makes no use of the model's predictions. Nevertheless, the method performs quite well on the third setup (SF + ID) when compared to the other methods that were trained with the greedy decision method, the least successful decision method of this setup.
Next, \textbf{\textit{MT-AVG}} performs well when the tasks are of equal importance (DP + NER), but achieves only moderate performance on the other setups.}

\guyfix{Surprisingly, \textbf{\textit{MT-MAX}} is highly effective despite its simplicity. It is mostly beneficial for the first two setups (DP + NER and NER + RE), where the tasks are of the same annotation granularity. It is the third best method overall, and it does not lag substantially behind the best method, \textit{MT-PAR}.
Interestingly, \textbf{\textit{MT-MIN}}, which offers a complementary perspective to \textit{MT-MAX}, is on average the worst MT-AL method, excluding \textit{MT-R}, and is mainly beneficial for the first setup (DP + NER).}

\guyfix{The next MT-AL method, \textbf{\textit{MT-PAR}}, seems to capture well the joint confidence space of the task pairs. It is the best method on average, achieving high average scores in all setups. However, when incorporating it with other training techniques, such as applying non-uniform weights (for the second setup), it is outperformed by the other MT-AL methods.}
\guyfix{\textbf{\textit{MT-RRF}} does not lag far behind \textit{MT-PAR}, achieving similar results on most tasks, excluding the RE and ID tasks, which are the higher-level tasks of their setups.}
\guyfix{Finally, \textbf{\textit{MT-IND}} does not excel in three of the four tasks of the first two setups, while achieving the best average results on NER, when jointly trained with RE. Furthermore, the method demonstrates strong performance on the third setup, when the tasks are of different annotation granularity, justifying the independent annotation selection in this case.}

\section{Conclusions}
\label{sec:conclusions}

We considered the problem of multi-task active learning for pre-trained Transformer-based models. We posed multiple research questions concerning the impact of multi-task modeling, multi-task selection criteria, overconfidence reduction, the relationships between the participating tasks, as well as budget constraints, and presented a systematic algorithmic and experimental investigation in order to answer these questions. Our results demonstrate the importance of MT-AL modeling in three challenging real-life scenarios, corresponding to diverse relations between the participating tasks.
In future work, we plan to research setups with more than two tasks and consider language generation and multilingual modeling.


\section{Acknowledgments}
\label{sec:ack}

We would like to thank the action editor and the reviewers, as well as the members of the IE@Technion NLP group for their valuable feedback and advice. This research was partially funded by an ISF personal grant No. 1625/18. RR also acknowledges the support of the Schmidt Career Advancement Chair in AI.

\bibliography{tacl2021}

\begin{thebibliography}{63}
\expandafter\ifx\csname natexlab\endcsname\relax\def\natexlab#1{#1}\fi

\bibitem[{Bai et~al.(2020)Bai, He, Liu, Zhao, and Nie}]{bai2020pre}
Guirong Bai, Shizhu He, Kang Liu, Jun Zhao, and Zaiqing Nie. 2020.
\newblock \href {https://aclanthology.org/2020.coling-main.130.pdf}
  {Pre-trained language model based active learning for sentence matching}.
\newblock In \emph{Proceedings of the 28th International Conference on
  Computational Linguistics}, pages 1495--1504.

\bibitem[{Chen et~al.(2018)Chen, Hou, Cheng, and Li}]{chen-etal-2018-joint}
Ying Chen, Wenjun Hou, Xiyao Cheng, and Shoushan Li. 2018.
\newblock \href {https://doi.org/10.18653/v1/D18-1066} {Joint learning for
  emotion classification and emotion cause detection}.
\newblock In \emph{Proceedings of the 2018 Conference on Empirical Methods in
  Natural Language Processing}, pages 646--651, Brussels, Belgium. Association
  for Computational Linguistics.

\bibitem[{Cormack et~al.(2009)Cormack, Clarke, and
  Buettcher}]{cormack2009reciprocal}
Gordon~V Cormack, Charles~LA Clarke, and Stefan Buettcher. 2009.
\newblock \href {https://dl.acm.org/doi/abs/10.1145/1571941.1572114}
  {Reciprocal rank fusion outperforms condorcet and individual rank learning
  methods}.
\newblock In \emph{Proceedings of the 32nd international ACM SIGIR conference
  on Research and development in information retrieval}, pages 758--759.

\bibitem[{Coucke et~al.(2018)Coucke, Saade, Ball, Bluche, Caulier, Leroy,
  Doumouro, Gisselbrecht, Caltagirone, Lavril, Primet, and
  Dureau}]{coucke2018snips}
Alice Coucke, Alaa Saade, Adrien Ball, Théodore Bluche, Alexandre Caulier,
  David Leroy, Clément Doumouro, Thibault Gisselbrecht, Francesco Caltagirone,
  Thibaut Lavril, Maël Primet, and Joseph Dureau. 2018.
\newblock \href {http://arxiv.org/abs/1805.10190} {Snips voice platform: an
  embedded spoken language understanding system for private-by-design voice
  interfaces}.
\newblock \emph{CoRR}, abs/1805.10190.

\bibitem[{Desai and Durrett(2020)}]{desai2020calibration}
Shrey Desai and Greg Durrett. 2020.
\newblock \href {https://www.aclweb.org/anthology/2020.emnlp-main.21.pdf}
  {Calibration of pre-trained transformers}.
\newblock In \emph{Proceedings of the 2020 Conference on Empirical Methods in
  Natural Language Processing (EMNLP)}, pages 295--302.

\bibitem[{Devlin et~al.(2019)Devlin, Chang, Lee, and
  Toutanova}]{devlin2019bert}
Jacob Devlin, Ming-Wei Chang, Kenton Lee, and Kristina Toutanova. 2019.
\newblock \href {https://www.aclweb.org/anthology/N19-1423.pdf} {Bert:
  Pre-training of deep bidirectional transformers for language understanding}.
\newblock In \emph{Proceedings of the 2019 Conference of the North American
  Chapter of the Association for Computational Linguistics: Human Language
  Technologies, Volume 1 (Long and Short Papers)}, pages 4171--4186.

\bibitem[{Dozat and Manning(2017)}]{dozat2017deep}
Timothy Dozat and Christopher~D. Manning. 2017.
\newblock \href
  {https://web.stanford.edu/~tdozat/files/TDozat-ICLR2017-Paper.pdf} {Deep
  biaffine attention for neural dependency parsing}.
\newblock In \emph{5th International Conference on Learning Representations,
  {ICLR} 2017, Toulon, France, April 24-26, 2017, Conference Track
  Proceedings}.

\bibitem[{Dror et~al.(2018)Dror, Baumer, Shlomov, and
  Reichart}]{dror2018hitchhiker}
Rotem Dror, Gili Baumer, Segev Shlomov, and Roi Reichart. 2018.
\newblock \href {https://aclanthology.org/P18-1128.pdf} {The hitchhiker’s
  guide to testing statistical significance in natural language processing}.
\newblock In \emph{Proceedings of the 56th Annual Meeting of the Association
  for Computational Linguistics (Volume 1: Long Papers)}, pages 1383--1392.

\bibitem[{Duong et~al.(2018)Duong, Afshar, Estival, Pink, Cohen, and
  Johnson}]{duong-etal-2018-active}
Long Duong, Hadi Afshar, Dominique Estival, Glen Pink, Philip Cohen, and Mark
  Johnson. 2018.
\newblock \href {https://doi.org/10.18653/v1/P18-2008} {Active learning for
  deep semantic parsing}.
\newblock In \emph{Proceedings of the 56th Annual Meeting of the Association
  for Computational Linguistics (Volume 2: Short Papers)}, pages 43--48,
  Melbourne, Australia. Association for Computational Linguistics.

\bibitem[{Ein-Dor et~al.(2020)Ein-Dor, Halfon, Gera, Shnarch, Dankin, Choshen,
  Danilevsky, Aharonov, Katz, and Slonim}]{ein-dor-etal-2020-active}
Liat Ein-Dor, Alon Halfon, Ariel Gera, Eyal Shnarch, Lena Dankin, Leshem
  Choshen, Marina Danilevsky, Ranit Aharonov, Yoav Katz, and Noam Slonim. 2020.
\newblock \href {https://doi.org/10.18653/v1/2020.emnlp-main.638} {{A}ctive
  {L}earning for {BERT}: {A}n {E}mpirical {S}tudy}.
\newblock In \emph{Proceedings of the 2020 Conference on Empirical Methods in
  Natural Language Processing (EMNLP)}, pages 7949--7962, Online. Association
  for Computational Linguistics.

\bibitem[{Finkel and Manning(2009)}]{finkel2009joint}
Jenny~Rose Finkel and Christopher~D Manning. 2009.
\newblock \href {https://aclanthology.org/N09-1037.pdf} {Joint parsing and
  named entity recognition}.
\newblock In \emph{Proceedings of Human Language Technologies: The 2009 Annual
  Conference of the North American Chapter of the Association for Computational
  Linguistics}, pages 326--334.

\bibitem[{Gal and Ghahramani(2016)}]{gal2016dropout}
Yarin Gal and Zoubin Ghahramani. 2016.
\newblock \href {http://proceedings.mlr.press/v48/gal16.pdf} {Dropout as a
  bayesian approximation: Representing model uncertainty in deep learning}.
\newblock In \emph{international conference on machine learning}, pages
  1050--1059. PMLR.

\bibitem[{Gardent et~al.(2017)Gardent, Shimorina, Narayan, and
  Perez-Beltrachini}]{gardent-etal-2017-webnlg}
Claire Gardent, Anastasia Shimorina, Shashi Narayan, and Laura
  Perez-Beltrachini. 2017.
\newblock \href {https://doi.org/10.18653/v1/W17-3518} {The {W}eb{NLG}
  challenge: Generating text from {RDF} data}.
\newblock In \emph{Proceedings of the 10th International Conference on Natural
  Language Generation}, pages 124--133, Santiago de Compostela, Spain.
  Association for Computational Linguistics.

\bibitem[{Grie{\ss}haber et~al.(2020)Grie{\ss}haber, Maucher, and
  Vu}]{griesshaber-etal-2020-fine}
Daniel Grie{\ss}haber, Johannes Maucher, and Ngoc~Thang Vu. 2020.
\newblock \href {https://doi.org/10.18653/v1/2020.coling-main.100} {Fine-tuning
  {BERT} for low-resource natural language understanding via active learning}.
\newblock In \emph{Proceedings of the 28th International Conference on
  Computational Linguistics}, pages 1158--1171, Barcelona, Spain (Online).
  International Committee on Computational Linguistics.

\bibitem[{Guo et~al.(2017)Guo, Pleiss, Sun, and
  Weinberger}]{guo2017calibration}
Chuan Guo, Geoff Pleiss, Yu~Sun, and Kilian~Q Weinberger. 2017.
\newblock \href {http://proceedings.mlr.press/v70/guo17a/guo17a.pdf} {On
  calibration of modern neural networks}.
\newblock In \emph{International Conference on Machine Learning}, pages
  1321--1330. PMLR.

\bibitem[{Haffari et~al.(2009)Haffari, Roy, and Sarkar}]{haffari2009active}
Gholamreza Haffari, Maxim Roy, and Anoop Sarkar. 2009.
\newblock \href {https://www.aclweb.org/anthology/N09-1047.pdf} {Active
  learning for statistical phrase-based machine translation}.
\newblock In \emph{Proceedings of Human Language Technologies: The 2009 Annual
  Conference of the North American Chapter of the Association for Computational
  Linguistics}, pages 415--423.

\bibitem[{Hochreiter and Schmidhuber(1997)}]{hochreiter1997long}
Sepp Hochreiter and J{\"u}rgen Schmidhuber. 1997.
\newblock \href
  {https://direct.mit.edu/neco/article/9/8/1735/6109/Long-Short-Term-Memory}
  {Long short-term memory}.
\newblock \emph{Neural computation}, 9(8):1735--1780.

\bibitem[{Hovy et~al.(2006)Hovy, Marcus, Palmer, Ramshaw, and
  Weischedel}]{hovy2006ontonotes}
Eduard Hovy, Mitch Marcus, Martha Palmer, Lance Ramshaw, and Ralph Weischedel.
  2006.
\newblock \href {https://www.aclweb.org/anthology/N06-2015.pdf} {Ontonotes: the
  90\% solution}.
\newblock In \emph{Proceedings of the human language technology conference of
  the NAACL, Companion Volume: Short Papers}, pages 57--60.

\bibitem[{Ikhwantri et~al.(2018)Ikhwantri, Louvan, Kurniawan, Abisena, Rachman,
  Wicaksono, and Mahendra}]{ikhwantri2018multi}
Fariz Ikhwantri, Samuel Louvan, Kemal Kurniawan, Bagas Abisena, Valdi Rachman,
  Alfan~Farizki Wicaksono, and Rahmad Mahendra. 2018.
\newblock \href {https://www.aclweb.org/anthology/W18-3406.pdf} {Multi-task
  active learning for neural semantic role labeling on low resource
  conversational corpus}.
\newblock In \emph{Proceedings of the Workshop on Deep Learning Approaches for
  Low-Resource NLP}, pages 43--50.

\bibitem[{Kim et~al.(2006)Kim, Song, Kim, Cha, and Lee}]{kim2006mmr}
Seokhwan Kim, Yu~Song, Kyungduk Kim, Jeong-Won Cha, and Gary~Geunbae Lee. 2006.
\newblock \href {https://aclanthology.org/N06-2018.pdf} {Mmr-based active
  machine learning for bio named entity recognition}.
\newblock In \emph{Proceedings of the Human Language Technology Conference of
  the NAACL, Companion Volume: Short Papers}, pages 69--72.

\bibitem[{Kingma and Ba(2015)}]{kingma2015adam}
Diederik~P Kingma and Jimmy Ba. 2015.
\newblock \href {https://arxiv.org/abs/1412.6980} {Adam: A method for
  stochastic optimization}.
\newblock In \emph{ICLR (Poster)}.

\bibitem[{Kong et~al.(2020)Kong, Jiang, Zhuang, Lyu, Zhao, and
  Zhang}]{kong-etal-2020-calibrated}
Lingkai Kong, Haoming Jiang, Yuchen Zhuang, Jie Lyu, Tuo Zhao, and Chao Zhang.
  2020.
\newblock \href {https://doi.org/10.18653/v1/2020.emnlp-main.102} {Calibrated
  language model fine-tuning for in- and out-of-distribution data}.
\newblock In \emph{Proceedings of the 2020 Conference on Empirical Methods in
  Natural Language Processing (EMNLP)}, pages 1326--1340, Online. Association
  for Computational Linguistics.

\bibitem[{Kulkarni et~al.(2018)Kulkarni, Xu, Ritter, and
  Machiraju}]{kulkarni-etal-2018-annotated}
Chaitanya Kulkarni, Wei Xu, Alan Ritter, and Raghu Machiraju. 2018.
\newblock \href {https://doi.org/10.18653/v1/N18-2016} {An annotated corpus for
  machine reading of instructions in wet lab protocols}.
\newblock In \emph{Proceedings of the 2018 Conference of the North {A}merican
  Chapter of the Association for Computational Linguistics: Human Language
  Technologies, Volume 2 (Short Papers)}, pages 97--106, New Orleans,
  Louisiana. Association for Computational Linguistics.

\bibitem[{Li et~al.(2020)Li, Stanovsky, and
  Zettlemoyer}]{li-etal-2020-active-learning}
Belinda~Z. Li, Gabriel Stanovsky, and Luke Zettlemoyer. 2020.
\newblock \href {https://doi.org/10.18653/v1/2020.acl-main.738} {Active
  learning for coreference resolution using discrete annotation}.
\newblock In \emph{Proceedings of the 58th Annual Meeting of the Association
  for Computational Linguistics}, pages 8320--8331, Online. Association for
  Computational Linguistics.

\bibitem[{Li et~al.(2021)Li, Ding, Shang, McAuley, and
  Feng}]{li-etal-2021-weakly}
Jiacheng Li, Haibo Ding, Jingbo Shang, Julian McAuley, and Zhe Feng. 2021.
\newblock \href {https://doi.org/10.18653/v1/2021.acl-long.352} {Weakly
  supervised named entity tagging with learnable logical rules}.
\newblock In \emph{Proceedings of the 59th Annual Meeting of the Association
  for Computational Linguistics and the 11th International Joint Conference on
  Natural Language Processing (Volume 1: Long Papers)}, pages 4568--4581,
  Online. Association for Computational Linguistics.

\bibitem[{Li et~al.(2016)Li, Zhang, Zhang, Liu, Chen, Wu, and
  Wang}]{li2016active}
Zhenghua Li, Min Zhang, Yue Zhang, Zhanyi Liu, Wenliang Chen, Hua Wu, and
  Haifeng Wang. 2016.
\newblock \href {https://www.aclweb.org/anthology/P16-1033.pdf} {Active
  learning for dependency parsing with partial annotation}.
\newblock In \emph{Proceedings of the 54th Annual Meeting of the Association
  for Computational Linguistics (Volume 1: Long Papers)}, pages 344--354.

\bibitem[{Lin et~al.(2018)Lin, Yang, Stoyanov, and Ji}]{lin2018multi}
Ying Lin, Shengqi Yang, Veselin Stoyanov, and Heng Ji. 2018.
\newblock \href {https://www.aclweb.org/anthology/P18-1074.pdf} {A
  multi-lingual multi-task architecture for low-resource sequence labeling}.
\newblock In \emph{Proceedings of the 56th Annual Meeting of the Association
  for Computational Linguistics (Volume 1: Long Papers)}, pages 799--809.

\bibitem[{Liu et~al.(2019{\natexlab{a}})Liu, He, Chen, and
  Gao}]{liu-etal-2019-multi-task}
Xiaodong Liu, Pengcheng He, Weizhu Chen, and Jianfeng Gao. 2019{\natexlab{a}}.
\newblock \href {https://doi.org/10.18653/v1/P19-1441} {Multi-task deep neural
  networks for natural language understanding}.
\newblock In \emph{Proceedings of the 57th Annual Meeting of the Association
  for Computational Linguistics}, pages 4487--4496, Florence, Italy.
  Association for Computational Linguistics.

\bibitem[{Liu et~al.(2019{\natexlab{b}})Liu, Ott, Goyal, Du, Joshi, Chen, Levy,
  Lewis, Zettlemoyer, and Stoyanov}]{liu2019roberta}
Yinhan Liu, Myle Ott, Naman Goyal, Jingfei Du, Mandar Joshi, Danqi Chen, Omer
  Levy, Mike Lewis, Luke Zettlemoyer, and Veselin Stoyanov. 2019{\natexlab{b}}.
\newblock \href {https://arxiv.org/pdf/1907.11692.pdf} {Roberta: A robustly
  optimized bert pretraining approach}.
\newblock \emph{arXiv preprint arXiv:1907.11692}.

\bibitem[{Lotov and Miettinen(2008)}]{lotov2008visualizing}
Alexander~V. Lotov and Kaisa Miettinen. 2008.
\newblock \href {https://link.springer.com/chapter/10.1007/978-3-540-88908-3_9}
  {Visualizing the pareto frontier}.
\newblock In \emph{Multiobjective optimization}, pages 213--243. Springer.

\bibitem[{Luan et~al.(2018)Luan, He, Ostendorf, and
  Hajishirzi}]{luan-etal-2018-multi}
Yi~Luan, Luheng He, Mari Ostendorf, and Hannaneh Hajishirzi. 2018.
\newblock \href {https://doi.org/10.18653/v1/D18-1360} {Multi-task
  identification of entities, relations, and coreference for scientific
  knowledge graph construction}.
\newblock In \emph{Proceedings of the 2018 Conference on Empirical Methods in
  Natural Language Processing}, pages 3219--3232, Brussels, Belgium.
  Association for Computational Linguistics.

\bibitem[{McCann et~al.(2017)McCann, Bradbury, Xiong, and
  Socher}]{mccann2017learned}
Bryan McCann, James Bradbury, Caiming Xiong, and Richard Socher. 2017.
\newblock \href {https://dl.acm.org/doi/pdf/10.5555/3295222.3295377} {Learned
  in translation: contextualized word vectors}.
\newblock In \emph{Proceedings of the 31st International Conference on Neural
  Information Processing Systems}, pages 6297--6308.

\bibitem[{Nayak and Ng(2020)}]{nayak2020effective}
Tapas Nayak and Hwee~Tou Ng. 2020.
\newblock \href {https://ojs.aaai.org/index.php/AAAI/article/view/6374}
  {Effective modeling of encoder-decoder architecture for joint entity and
  relation extraction}.
\newblock In \emph{Proceedings of the AAAI conference on artificial
  intelligence}, volume~34, pages 8528--8535.

\bibitem[{Nguyen and Nguyen(2021)}]{nguyen-nguyen-2021-phonlp}
Linh~The Nguyen and Dat~Quoc Nguyen. 2021.
\newblock \href {https://www.aclweb.org/anthology/2021.naacl-demos.1}
  {{P}ho{NLP}: A joint multi-task learning model for {V}ietnamese
  part-of-speech tagging, named entity recognition and dependency parsing}.
\newblock In \emph{Proceedings of the 2021 Conference of the North American
  Chapter of the Association for Computational Linguistics: Human Language
  Technologies: Demonstrations}, pages 1--7, Online. Association for
  Computational Linguistics.

\bibitem[{Nivre et~al.(2020)Nivre, de~Marneffe, Ginter, Haji{\v{c}}, Manning,
  Pyysalo, Schuster, Tyers, and Zeman}]{nivre2020universal}
Joakim Nivre, Marie-Catherine de~Marneffe, Filip Ginter, Jan Haji{\v{c}},
  Christopher~D. Manning, Sampo Pyysalo, Sebastian Schuster, Francis Tyers, and
  Daniel Zeman. 2020.
\newblock \href {https://aclanthology.org/2020.lrec-1.497} {{U}niversal
  {D}ependencies v2: An evergrowing multilingual treebank collection}.
\newblock In \emph{Proceedings of the 12th Language Resources and Evaluation
  Conference}, pages 4034--4043, Marseille, France. European Language Resources
  Association.

\bibitem[{Ovadia et~al.(2019)Ovadia, Fertig, Ren, Nado, Sculley, Nowozin,
  Dillon, Lakshminarayanan, and Snoek}]{ovadia2019can}
Yaniv Ovadia, Emily Fertig, Jie Ren, Zachary Nado, D~Sculley, Sebastian
  Nowozin, Joshua Dillon, Balaji Lakshminarayanan, and Jasper Snoek. 2019.
\newblock \href
  {https://proceedings.neurips.cc/paper/2019/file/8558cb408c1d76621371888657d2eb1d-Paper.pdf}
  {Can you trust your model's uncertainty? evaluating predictive uncertainty
  under dataset shift}.
\newblock \emph{Advances in Neural Information Processing Systems},
  32:13991--14002.

\bibitem[{Peris and Casacuberta(2018)}]{peris-casacuberta-2018-active}
{\'A}lvaro Peris and Francisco Casacuberta. 2018.
\newblock \href {https://doi.org/10.18653/v1/K18-1015} {Active learning for
  interactive neural machine translation of data streams}.
\newblock In \emph{Proceedings of the 22nd Conference on Computational Natural
  Language Learning}, pages 151--160, Brussels, Belgium. Association for
  Computational Linguistics.

\bibitem[{Peters et~al.(2018)Peters, Neumann, Iyyer, Gardner, Clark, Lee, and
  Zettlemoyer}]{peters2018deep}
Matthew Peters, Mark Neumann, Mohit Iyyer, Matt Gardner, Christopher Clark,
  Kenton Lee, and Luke Zettlemoyer. 2018.
\newblock \href {https://www.aclweb.org/anthology/N18-1202.pdf} {Deep
  contextualized word representations}.
\newblock In \emph{Proceedings of the 2018 Conference of the North American
  Chapter of the Association for Computational Linguistics: Human Language
  Technologies, Volume 1 (Long Papers)}, pages 2227--2237.

\bibitem[{Price(1990)}]{price1990evaluation}
Patti Price. 1990.
\newblock \href {https://aclanthology.org/H90-1020.pdf} {Evaluation of spoken
  language systems: The atis domain}.
\newblock In \emph{Speech and Natural Language: Proceedings of a Workshop Held
  at Hidden Valley, Pennsylvania, June 24-27, 1990}.

\bibitem[{Raffel et~al.(2020)Raffel, Shazeer, Roberts, Lee, Narang, Matena,
  Zhou, Li, and Liu}]{raffel2020exploring}
Colin Raffel, Noam Shazeer, Adam Roberts, Katherine Lee, Sharan Narang, Michael
  Matena, Yanqi Zhou, Wei Li, and Peter~J Liu. 2020.
\newblock \href {https://m.jmlr.org/papers/volume21/20-074/20-074.pdf}
  {Exploring the limits of transfer learning with a unified text-to-text
  transformer}.
\newblock \emph{Journal of Machine Learning Research}, 21:1--67.

\bibitem[{Reichart and Rappoport(2007)}]{reichart2007ensemble}
Roi Reichart and Ari Rappoport. 2007.
\newblock \href {https://www.aclweb.org/anthology/P07-1052.pdf} {An ensemble
  method for selection of high quality parses}.
\newblock In \emph{Proceedings of the 45th Annual Meeting of the Association of
  Computational Linguistics}, pages 408--415.

\bibitem[{Reichart and Rappoport(2009)}]{reichart2009sample}
Roi Reichart and Ari Rappoport. 2009.
\newblock \href {https://aclanthology.org/W09-1103.pdf} {Sample selection for
  statistical parsers: Cognitively driven algorithms and evaluation measures}.
\newblock In \emph{Proceedings of the Thirteenth Conference on Computational
  Natural Language Learning (CoNLL-2009)}, pages 3--11.

\bibitem[{Reichart et~al.(2008)Reichart, Tomanek, Hahn, and
  Rappoport}]{reichart2008multi}
Roi Reichart, Katrin Tomanek, Udo Hahn, and Ari Rappoport. 2008.
\newblock \href {https://www.aclweb.org/anthology/P08-1098.pdf} {Multi-task
  active learning for linguistic annotations}.
\newblock In \emph{Proceedings of ACL-08: HLT}, pages 861--869.

\bibitem[{Rotman and Reichart(2019)}]{rotman2019deep}
Guy Rotman and Roi Reichart. 2019.
\newblock \href
  {https://watermark.silverchair.com/tacl_a_00294.pdf?token=AQECAHi208BE49Ooan9kkhW_Ercy7Dm3ZL_9Cf3qfKAc485ysgAAAtAwggLMBgkqhkiG9w0BBwagggK9MIICuQIBADCCArIGCSqGSIb3DQEHATAeBglghkgBZQMEAS4wEQQMzpI7Jr-lNdwiehweAgEQgIICgwQvHOhvqKn8SbMVhgqIEelusTnBVTmSxheMxxjOwULyV5RKwPo0ZXvKuBVviVW6ncAmnLgllfCh09Md9VBNUe1jCeQKdf5oWz_ofW3giERyQBEnVuKEsqZhTXU43aqyQxfQgNgKfz1YFU8XysFkuXRzfseekaOcyyCVhwsA8tY3Yc8K8UByk4TuizM5-vP5O6TNDPtCA_pyk-1p_7YtdMWA-uum9bQeNzf-JKh1gH1jyu64yvi5gByL6xk4VMQypJfK4KvnDYoKeEsBbusMw_KI-CrcKwe3NA7j2n4Y9iaR-M9iFeCv1fCOIRetri9Wdo5O0wndXtIilXMlqyWKqR0d_u73BP69BQoliNSLVfeYU-Bjw7Dz_-a_MxSWqKzVF_MenkRZyZxVxsAPfkYf8nakbtbvuNi_g3bEk-uWnP7WoPiOE-nQnDu1prLLgCD6D9wmIRdxYzdZjSaGezgPO871knRb6frw6ozu1irIP0EdlmG1f-41xFzNMiqf49czRHhHxMfUFnWDlfDN75r7BGwM77TXQsYOWzh7yv2dYDfNs4OnepBeKzdmRLDGOqQ6ciEEdP-qqudslQqIn_sXMKhuIFBuoNqYzNm1b67tJlabFOB7WqSl-wFuxzymneBgfj6qQTrIQGf3XYBFormTAxzw20zZWheygXsoE-JW-4Bw9xUVl1MK9EBvP6KE-8kuYXWBzqlWllQuQ2aJBz1XQZQVjc1TK_JqwFev8Drzd6vmtV2JuswEYyoWr-qkYjru61DZh_KU-mH4JT3S7AKDOpNtLCulBZ4c2zQPDAPkf6H7OCYWYrDO2vC-j-ECTQz4YPBJmoWqdMzMRYprx4SQIGixrWg}
  {Deep contextualized self-training for low resource dependency parsing}.
\newblock \emph{Transactions of the Association for Computational Linguistics},
  7:695--713.

\bibitem[{Ruder(2017)}]{ruder2017overview}
Sebastian Ruder. 2017.
\newblock \href {https://arxiv.org/abs/1706.05098} {An overview of multi-task
  learning in deep neural networks}.
\newblock \emph{arXiv preprint arXiv:1706.05098}.

\bibitem[{Safi~Samghabadi et~al.(2020)Safi~Samghabadi, Patwa, PYKL, Mukherjee,
  Das, and Solorio}]{safi-samghabadi-etal-2020-aggression}
Niloofar Safi~Samghabadi, Parth Patwa, Srinivas PYKL, Prerana Mukherjee,
  Amitava Das, and Thamar Solorio. 2020.
\newblock \href {https://www.aclweb.org/anthology/2020.trac-1.20} {Aggression
  and misogyny detection using {BERT}: A multi-task approach}.
\newblock In \emph{Proceedings of the Second Workshop on Trolling, Aggression
  and Cyberbullying}, pages 126--131, Marseille, France. European Language
  Resources Association (ELRA).

\bibitem[{Sanh et~al.(2019)Sanh, Wolf, and Ruder}]{sanh2019hierarchical}
Victor Sanh, Thomas Wolf, and Sebastian Ruder. 2019.
\newblock \href {https://www.aaai.org/ojs/index.php/AAAI/article/view/4673} {A
  hierarchical multi-task approach for learning embeddings from semantic
  tasks}.
\newblock In \emph{Proceedings of the AAAI Conference on Artificial
  Intelligence}, volume~33, pages 6949--6956.

\bibitem[{Schneider et~al.(2018)Schneider, Hwang, Srikumar, Prange, Blodgett,
  Moeller, Stern, Bitan, and Abend}]{schneider2018comprehensive}
Nathan Schneider, Jena~D. Hwang, Vivek Srikumar, Jakob Prange, Austin Blodgett,
  Sarah~R. Moeller, Aviram Stern, Adi Bitan, and Omri Abend. 2018.
\newblock \href {https://doi.org/10.18653/v1/P18-1018} {Comprehensive
  supersense disambiguation of {E}nglish prepositions and possessives}.
\newblock In \emph{Proceedings of the 56th Annual Meeting of the Association
  for Computational Linguistics (Volume 1: Long Papers)}, pages 185--196,
  Melbourne, Australia. Association for Computational Linguistics.

\bibitem[{Settles and Craven(2008)}]{settles2008analysis}
Burr Settles and Mark Craven. 2008.
\newblock \href {https://www.aclweb.org/anthology/D08-1112.pdf} {An analysis of
  active learning strategies for sequence labeling tasks}.
\newblock In \emph{Proceedings of the 2008 Conference on Empirical Methods in
  Natural Language Processing}, pages 1070--1079.

\bibitem[{Seung et~al.(1992)Seung, Opper, and Sompolinsky}]{seung1992query}
H~Sebastian Seung, Manfred Opper, and Haim Sompolinsky. 1992.
\newblock \href {https://dl.acm.org/doi/pdf/10.1145/130385.130417} {Query by
  committee}.
\newblock In \emph{Proceedings of the fifth annual workshop on Computational
  learning theory}, pages 287--294.

\bibitem[{Shen et~al.(2017)Shen, Yun, Lipton, Kronrod, and
  Anandkumar}]{shen-etal-2017-deep}
Yanyao Shen, Hyokun Yun, Zachary Lipton, Yakov Kronrod, and Animashree
  Anandkumar. 2017.
\newblock \href {https://doi.org/10.18653/v1/W17-2630} {Deep active learning
  for named entity recognition}.
\newblock In \emph{Proceedings of the 2nd Workshop on Representation Learning
  for {NLP}}, pages 252--256, Vancouver, Canada. Association for Computational
  Linguistics.

\bibitem[{S{\o}gaard and Goldberg(2016)}]{sogaard2016deep}
Anders S{\o}gaard and Yoav Goldberg. 2016.
\newblock \href {https://aclanthology.org/P16-2038.pdf} {Deep multi-task
  learning with low level tasks supervised at lower layers}.
\newblock In \emph{Proceedings of the 54th Annual Meeting of the Association
  for Computational Linguistics (Volume 2: Short Papers)}, pages 231--235.

\bibitem[{Szegedy et~al.(2016)Szegedy, Vanhoucke, Ioffe, Shlens, and
  Wojna}]{szegedy2016rethinking}
Christian Szegedy, Vincent Vanhoucke, Sergey Ioffe, Jon Shlens, and Zbigniew
  Wojna. 2016.
\newblock \href
  {https://www.cv-foundation.org/openaccess/content_cvpr_2016/papers/Szegedy_Rethinking_the_Inception_CVPR_2016_paper.pdf}
  {Rethinking the inception architecture for computer vision}.
\newblock In \emph{Proceedings of the IEEE conference on computer vision and
  pattern recognition}, pages 2818--2826.

\bibitem[{Thulasidasan et~al.(2019)Thulasidasan, Chennupati, Bilmes,
  Bhattacharya, and Michalak}]{sunilmixup2019}
Sunil Thulasidasan, Gopinath Chennupati, Jeff~A Bilmes, Tanmoy Bhattacharya,
  and Sarah Michalak. 2019.
\newblock \href
  {https://proceedings.neurips.cc/paper/2019/file/36ad8b5f42db492827016448975cc22d-Paper.pdf}
  {On mixup training: Improved calibration and predictive uncertainty for deep
  neural networks}.
\newblock In \emph{Advances in Neural Information Processing Systems},
  volume~32. Curran Associates, Inc.

\bibitem[{Tomanek and Hahn(2010)}]{tomanek2010comparison}
Katrin Tomanek and Udo Hahn. 2010.
\newblock \href {https://aclanthology.org/C10-2143.pdf} {A comparison of models
  for cost-sensitive active learning}.
\newblock In \emph{Coling 2010: Posters}, pages 1247--1255.

\bibitem[{Vaswani et~al.(2017)Vaswani, Shazeer, Parmar, Uszkoreit, Jones,
  Gomez, Kaiser, and Polosukhin}]{vaswani2017attention}
Ashish Vaswani, Noam Shazeer, Niki Parmar, Jakob Uszkoreit, Llion Jones,
  Aidan~N Gomez, {\L}ukasz Kaiser, and Illia Polosukhin. 2017.
\newblock \href
  {https://papers.nips.cc/paper/7181-attention-is-all-you-need.pdf} {Attention
  is all you need}.
\newblock In \emph{Advances in neural information processing systems}, pages
  5998--6008.

\bibitem[{Wang et~al.(2018)Wang, Singh, Michael, Hill, Levy, and
  Bowman}]{wang-etal-2018-glue}
Alex Wang, Amanpreet Singh, Julian Michael, Felix Hill, Omer Levy, and Samuel
  Bowman. 2018.
\newblock \href {https://doi.org/10.18653/v1/W18-5446} {{GLUE}: A multi-task
  benchmark and analysis platform for natural language understanding}.
\newblock In \emph{Proceedings of the 2018 {EMNLP} Workshop {B}lackbox{NLP}:
  Analyzing and Interpreting Neural Networks for {NLP}}, pages 353--355,
  Brussels, Belgium. Association for Computational Linguistics.

\bibitem[{Wiatrak and Iso-Sipila(2020)}]{wiatrak-iso-sipila-2020-simple}
Maciej Wiatrak and Juha Iso-Sipila. 2020.
\newblock \href {https://doi.org/10.18653/v1/2020.louhi-1.2} {Simple
  hierarchical multi-task neural end-to-end entity linking for biomedical
  text}.
\newblock In \emph{Proceedings of the 11th International Workshop on Health
  Text Mining and Information Analysis}, pages 12--17, Online. Association for
  Computational Linguistics.

\bibitem[{Wolf et~al.(2020)Wolf, Debut, Sanh, Chaumond, Delangue, Moi, Cistac,
  Rault, Louf, Funtowicz, Davison, Shleifer, von Platen, Ma, Jernite, Plu, Xu,
  Scao, Gugger, Drame, Lhoest, and Rush}]{wolf2019huggingface}
Thomas Wolf, Lysandre Debut, Victor Sanh, Julien Chaumond, Clement Delangue,
  Anthony Moi, Pierric Cistac, Tim Rault, Rémi Louf, Morgan Funtowicz, Joe
  Davison, Sam Shleifer, Patrick von Platen, Clara Ma, Yacine Jernite, Julien
  Plu, Canwen Xu, Teven~Le Scao, Sylvain Gugger, Mariama Drame, Quentin Lhoest,
  and Alexander~M. Rush. 2020.
\newblock \href {https://www.aclweb.org/anthology/2020.emnlp-demos.6}
  {Transformers: State-of-the-art natural language processing}.
\newblock In \emph{Proceedings of the 2020 Conference on Empirical Methods in
  Natural Language Processing: System Demonstrations}, pages 38--45.
  Association for Computational Linguistics.

\bibitem[{Xie et~al.(2018)Xie, Chang, Ren, Chen, and Yu}]{xie-etal-2018-cost}
Kaige Xie, Cheng Chang, Liliang Ren, Lu~Chen, and Kai Yu. 2018.
\newblock \href {https://doi.org/10.18653/v1/W18-5022} {Cost-sensitive active
  learning for dialogue state tracking}.
\newblock In \emph{Proceedings of the 19th Annual {SIG}dial Meeting on
  Discourse and Dialogue}, pages 209--213, Melbourne, Australia. Association
  for Computational Linguistics.

\bibitem[{Zhao et~al.(2020)Zhao, Huang, Zhang, Lu, and
  Xue}]{zhao-etal-2020-spanmlt}
He~Zhao, Longtao Huang, Rong Zhang, Quan Lu, and Hui Xue. 2020.
\newblock \href {https://doi.org/10.18653/v1/2020.acl-main.296} {{S}pan{M}lt: A
  span-based multi-task learning framework for pair-wise aspect and opinion
  terms extraction}.
\newblock In \emph{Proceedings of the 58th Annual Meeting of the Association
  for Computational Linguistics}, pages 3239--3248, Online. Association for
  Computational Linguistics.

\bibitem[{Zhu et~al.(2020)Zhu, Ye, Luo, and Zhang}]{zhu-etal-2020-multitask}
Hua Zhu, Wu~Ye, Sihan Luo, and Xidong Zhang. 2020.
\newblock \href {https://doi.org/10.18653/v1/2020.coling-main.430} {A multitask
  active learning framework for natural language understanding}.
\newblock In \emph{Proceedings of the 28th International Conference on
  Computational Linguistics}, pages 4900--4914, Barcelona, Spain (Online).
  International Committee on Computational Linguistics.

\bibitem[{Zhu and Yu(2017)}]{zhu2017encoder}
Su~Zhu and Kai Yu. 2017.
\newblock \href {https://ieeexplore.ieee.org/stamp/stamp.jsp?arnumber=7953243}
  {Encoder-decoder with focus-mechanism for sequence labelling based spoken
  language understanding}.
\newblock In \emph{2017 IEEE International Conference on Acoustics, Speech and
  Signal Processing (ICASSP)}, pages 5675--5679. IEEE.

\end{thebibliography}
\bibliographystyle{acl_natbib}

\end{document}